%% file: bare_conf_compsoc.tex
\documentclass[conference,compsoc]{IEEEtran}
\usepackage{graphicx}
\usepackage{booktabs}
\usepackage{multirow}
\usepackage{amsfonts}
\usepackage{amsmath}
\usepackage[ruled,vlined]{algorithm2e}
\usepackage{float}
\usepackage{afterpage}
\usepackage{graphicx}
\usepackage{subcaption} 
\usepackage{xcolor} 
\usepackage{tcolorbox}

\definecolor{L_blue}{RGB}{31, 119, 180}
\definecolor{D_blue}{RGB}{0, 51, 160}
\definecolor{D_orange}{RGB}{255, 127, 14}
\definecolor{D_red}{RGB}{214, 39, 40}

\ifCLASSOPTIONcompsoc
  \usepackage[nocompress]{cite}
\else
  \usepackage{cite}
\fi
\ifCLASSINFOpdf
\else
\fi
\begin{document}
%
% paper title
% Titles are generally capitalized except for words such as a, an, and, as,
% at, but, by, for, in, nor, of, on, or, the, to and up, which are usually
% not capitalized unless they are the first or last word of the title.
% Linebreaks \\ can be used within to get better formatting as desired.
% Do not put math or special symbols in the title.
\title{AttenMIA: LLM Membership Inference Attack through Attention Signals}

% author names and affiliations
% use a multiple-column layout for up to three different
% affiliations

% Add authors here

\author{
\textbf{Pedram Zaree}$^{1}$ \quad
\textbf{Md Abdullah Al Mamun}$^{1}$ \quad
\textbf{Yue Dong}$^{1}$ \\
\textbf{Ihsen Alouani}$^{2}$ \quad
\textbf{Nael Abu-Ghazaleh}$^{1}$ \\[1.5ex]
$^{1}$CSE Department,  
University of California, Riverside, USA \\  
$^{2}$CSIT, Queen’s University Belfast, UK \\[1.0ex]
\texttt{\{pzare003, mmamu003, yued, naelag\}@ucr.edu} \quad
\texttt{i.alouani@qub.ac.uk}
}

\maketitle

\input{sections/abstract}

% For peer review papers, you can put extra information on the cover
% page as needed:
% \ifCLASSOPTIONpeerreview
% \begin{center} \bfseries EDICS Category: 3-BBND \end{center}
% \fi
%
% For peerreview papers, this IEEEtran command inserts a page break and
% creates the second title. It will be ignored for other modes.
\IEEEpeerreviewmaketitle

\input{sections/introduction}

\input{sections/background}

\input{sections/threat_model}

\input{sections/methodology}

\input{sections/experiments}

\input{sections/data_extraction}
\input{sections/related_work}

\input{sections/conclusion}
\input{sections/acknowledgments}

% References
\bibliographystyle{IEEEtran}
\bibliography{reference}

% Appendix
\input{sections/appendix}

\end{document}

%% file: sections/abstract.tex
% As a general rule, do not put math, special symbols, or citations
% in the abstract

\begin{abstract}

Large Language Models (LLMs) are increasingly deployed to enable or improve a multitude of real-world applications.  Given the large size of their training data sets, their tendency to memorize training data raises serious privacy and intellectual property concerns. A key threat is the \textit{membership inference attack} (MIA), which aims to determine whether a given sample was included in the model’s training set. Existing MIAs for LLMs rely primarily on output confidence scores or embedding-based features, but these signals are often brittle, leading to limited attack success. 
We introduce \textbf{AttenMIA}, a new MIA framework that exploits self-attention patterns inside the transformer model to infer membership. Attention controls the information flow within the transformer, exposing different patterns for memorization that can be used to identify members of the dataset.  Our method uses information from attention heads across layers and combines them with perturbation-based divergence metrics to train an effective MIA classifier. 
Using extensive experiments on open-source models including LLaMA-2, Pythia, and Opt models, we show that attention-based features consistently outperform baselines, particularly under the important low-false-positive metric (e.g., achieving up to \textbf{0.996 ROC AUC} \& \textbf{87.9\% TPR@1\%FPR} on the WikiMIA-32 benchmark with Llama2-13b).  We show that attention signals generalize across datasets and architectures, and provide a layer- and head-level analysis of where membership leakage is most pronounced.   We also show that using \textbf{AttenMIA} to replace other membership inference attacks in a data extraction framework results in training data extraction attacks that outperform the state of the art.  Our findings reveal that attention mechanisms, originally introduced to enhance interpretability, can inadvertently amplify privacy risks in LLMs, underscoring the need for new defenses.

\end{abstract}

%% file: sections/introduction.tex
\section{Introduction}

% Explain the LLMs and memorization case as a privacy risk
Large Language Models (LLMs) have become essential components integrated into diverse applications, from conversational assistants \cite{abdullah2022chatgpt} to coding assistants \cite{wang2023review} to application domains such as biomedical discovery \cite{lu2024large}. LLM's impressive performance and diverse learning capabilities stem from training on massive corpora supporting generalization across different tasks and domains.  However, recent work has shown that LLMs can \emph{memorize} training data~\cite{carlini2021extracting,kandpal2022deduplicating,al2023deepmem}, raising serious privacy and security concerns. For instance, Carlini et al.~\cite{carlini2021extracting} demonstrated large-scale extraction of verbatim training examples from GPT models, and Kandpal et al.~\cite{kandpal2022deduplicating} reported leakage of personally identifiable information. Such memorization poses a threat to both individual privacy and the intellectual property of data used in these datasets.

% Introduce MIA as a case of memorization
One of the important attacks targeting datasets is the \emph{membership inference attack} (MIA) \cite{hu2022membership}, which seeks to determine whether a given sample was part of a model’s training data \cite{shokri2017membership}. Successful MIAs undermine data confidentiality and can serve as building blocks for more powerful data extraction attacks that incrementally recover the data.  MIAs rely on the observation that machine learning models overfit to their training data, exposing membership through their internal state or output (e.g., as higher confidence/lower perplexity).  %In the context of LLMs, MIAs have been applied both at the pre-training stage---e.g., via perplexity thresholding \cite{carlini2021extracting, mattern2023membership}---and at the fine-tuning stage, where overfitting amplifies leakage \cite{mireshghallah2022empirical, fu2024membership}. 

% Explain baselines and their weaknesses
Most existing MIAs for LLMs rely on output-based signals such as log-likelihoods, perplexity, or token-level confidence scores \cite{yeom2018privacy, carlini2021extracting, mattern2023membership, xie2024recall}. While effective in some cases, these methods suffer from two key limitations. First, their decision boundaries often exhibit significant overlap between members and non-members, resulting in degraded performance (for membership inference, often performance under low false-positive thresholds is most important). Second, they are brittle under distribution shifts, as output statistics do not always generalize when evaluation data distributions differ from the training data distributions. More advanced approaches, such as LiRA \cite{carlini2022membership} or semantic MIAs \cite{mozaffari2024semantic}, improve accuracy but remain computationally expensive and limited in interpretability.

% Explain that attention has not been explored for this purpose
Beyond MIAs that exploit output-based signals, feature-based MIAs exploit internal model states, such as gradients \cite{nasr2019comprehensive} or hidden representations \cite{song2019privacy} to separate member from non-member samples. Despite the centrality of attention mechanisms in transformer architectures~\cite{vaswani2017attention,dong2021attention,ben2024attend}, attention has not been systematically studied as a source of membership leakage.  Given the critical role that attention plays in information flow in transformers, and its uses in interpretability, there is reason to believe that it may provide discriminative features with respect to membership inference.
%This gap is especially striking given that attention was originally introduced to improve interpretability \cite{vaswani2017attention}, but may itself encode memorization artifacts.

 \begin{figure*}[t!]
\centering
\includegraphics[width=0.95\linewidth]{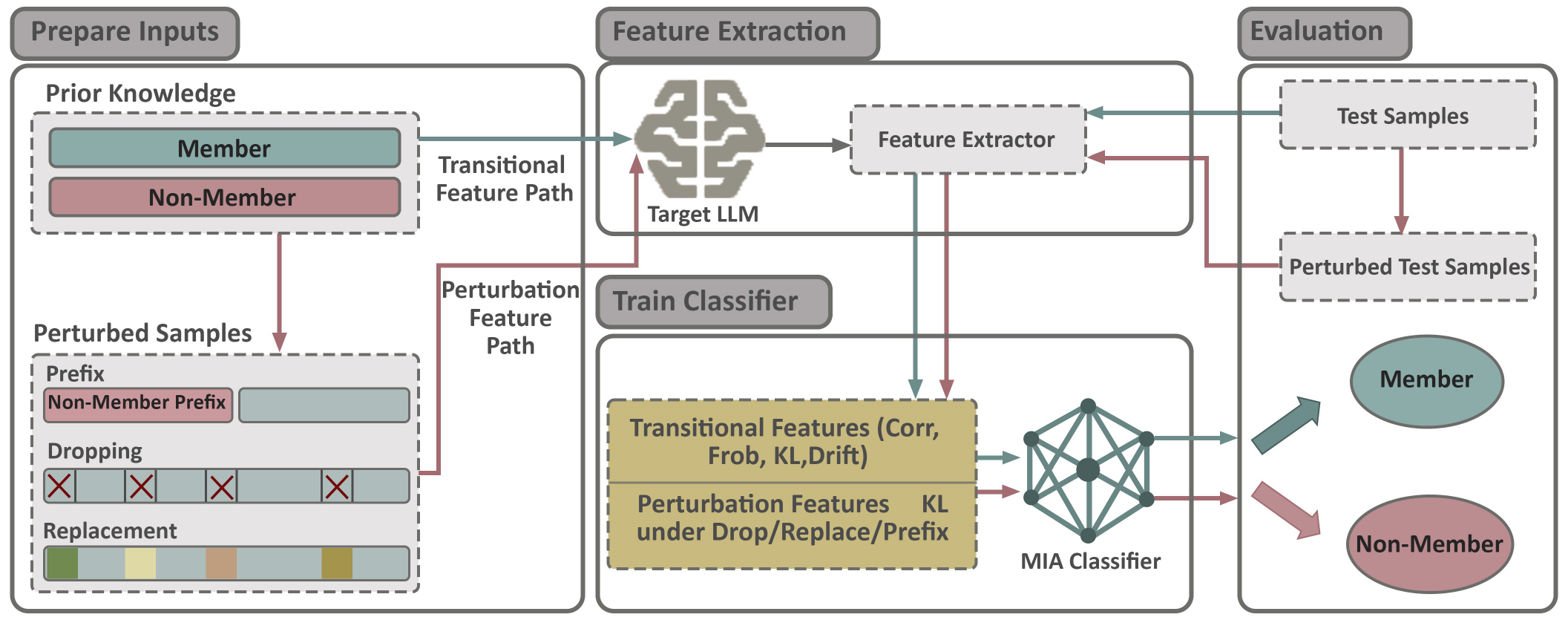}
\caption{Overview of \textsc{AttenMIA}, our attention-based membership inference framework. 
The pipeline consists of two parallel feature pathways: 
(1) \emph{Transitional features}, which quantify consistency and drift of attention patterns across layers and heads (e.g., correlation, Frobenius distance, KL divergence, barycenter drift), 
and (2) \emph{Perturbation features}, which capture how attention distributions shift under controlled input modifications (token dropping, replacement, prefix insertion). 
Both feature families are extracted from the target LLM and concatenated into a feature vector $\mathbf{z}$, which is used to train an MIA classifier. 
At evaluation time, the classifier outputs a membership probability to decide whether a target sample was part of the training data.}

\label{fig: MIA_framework}
\end{figure*}

% Shows our contribution
\textbf{Our work:} In this work, we introduce \textsc{AttenMIA}, the first attention-based white-box MIA framework targeting LLMs. We hypothesize that pretraining induces more structured and concentrated attention patterns for member samples, and these signals persist across layers and heads, providing opportunities for identifying members of the training data set. To test this hypothesis, we study a number of attention parameter-based features that capture inter-layer attention patterns. We show that these features enable highly effective membership inference, significantly outperforming state-of-the-art models based on output confidence or model parameter features. % without relying on output logits or reference models.

Beyond simply replacing output statistics with attention values, \textsc{AttenMIA} leverages the unique multi-head structure of transformer models to track and identify fine-grained privacy signals based on rate of change across the layers. We introduce layer-wise and head-wise correlation measures and barycentric drift metrics.  We also consider active attacks where we perturb the input sample and observe resulting changes in the attention metrics, with the intuition that perturbing member samples (turning them into non-members) results in larger shifts than perturbing non-member samples (which then remain non-members).   This approach captures both the stability and expressivity of a model’s internal attention flows, providing a more faithful representation of memorization than previously explored features.  

An overview of our framework is shown in Figure \ref{fig: MIA_framework}, where we extract transitional and perturbation-based attention features from the target model and use them to train an MIA classifier. We demonstrate that the derived features are robust across architectures and scales. By analyzing models including LLaMA-2, Pythia, and OPT, we show that membership cues encoded in attention persist even under model variations and moderate domain shifts. This robustness suggests that attention-based MIAs can generalize more effectively than output-based or gradient-based methods, which often require retraining or reference data. Through extensive experiments on open-weight models across multiple benchmarks (WikiMIA, MIMIR), we demonstrate that attention-derived features consistently outperform output-based baselines. Notably, our method achieves up to \textbf{0.996 ROC AUC} \& \textbf{87.9\% TPR@1\%FPR} on the WikiMIA-32 benchmark with Llama2-13b, setting a new state of the art in the critical low-false-positive regime.  We also show that integrating \textsc{AttenMIA} in a training data extraction framework (replacing a state of the art MIA algorithm) results in substantially higher data extraction success.   Beyond performance, we provide a fine-grained analysis of which layers and heads encode the strongest membership signals, revealing that attention mechanisms inadvertently amplify privacy risks in LLMs.

%% file: sections/background.tex
\section{Background}

In this section, we present the theoretical foundations that motivate our attack. We first review membership inference in machine learning, emphasizing its evolution in the context of large language models (LLMs). We then introduce the attention mechanism in transformers and discuss why its internal dynamics may act as a carrier of privacy-relevant information.

\subsection{Membership Inference in Machine Learning}
Membership inference attacks (MIAs) aim to answer a fundamental privacy question: \emph{given a trained model and a data sample, can an adversary determine whether the sample was part of the model’s training set?} The possibility of such inference arises because learning algorithms often behave differently on examples they have seen during training compared to those they have not. In overparameterized models, this distinction can emerge from subtle differences in prediction confidence, gradient norms, or internal representations, even when the model generalizes well overall.

Formally, for a model $f_{\theta}$ and a labeled example $(x, y)$, a membership inference attack seeks to estimate a binary variable $m(x, y) \in \{0,1\}$ such that $m(x, y) = 1$ if and only if $(x, y)$ was included in the training set. Early approaches exploited output-level statistics, such as classification confidence or loss values, to distinguish members from non-members~\cite{shokri2017membership, yeom2018privacy}. Subsequent work revealed that this vulnerability is not confined to outputs: intermediate representations within the model can also leak membership information.

Large language models (LLMs) amplify this challenge. Unlike classifiers that produce a single probability vector, LLMs generate text token by token, distributing uncertainty across long sequences. Their internal activations encode contextual dependencies rather than direct confidence scores. As a result, membership inference in LLMs often requires probing deeper into the model’s hidden states and dynamics. In particular, attention distributions, central to how transformers contextualize information, may carry unique and stable signals that differentiate training samples from unseen data.

This shift from output-based to representation-based inference motivates exploring \emph{where} and \emph{how} membership information is embedded within LLMs. Understanding these internal privacy signals forms the foundation for the attention-driven analysis developed in our proposed framework, \textbf{AttenMIA}.

\subsection{Attention Mechanisms in LLMs}

Transformer architectures~\cite{vaswani2017attention} form the foundation of most contemporary LLMs, with self-attention enabling each token to contextualize its representation by attending to other tokens in the sequence. At a given layer $\ell$, hidden states $h^{(\ell)} \in \mathbb{R}^{L \times d}$ are projected into query, key, and value spaces:
\begin{equation}
Q = h^{(\ell)} W_Q, \quad K = h^{(\ell)} W_K, \quad V = h^{(\ell)} W_V,
\end{equation}
where $W_Q, W_K, W_V \in \mathbb{R}^{d \times d_h}$ are learned parameters. The attention matrix is computed as
\begin{equation}
A = \mathrm{softmax}\!\left(\tfrac{QK^{\top}}{\sqrt{d_h}}\right),
\end{equation}
and the contextualized output is
\begin{equation}
O = A V.
\end{equation}
In multi-head attention (MHA), $H$ heads operate in parallel to capture diverse relational structures:
\begin{equation}
\mathrm{MHA}(X) = \mathrm{Concat}(O^{(1)}, \ldots, O^{(H)})W_O.
\end{equation}

Each head learns to emphasize different aspects of linguistic or structural dependency, syntax, coreference, or long-range semantics. The resulting attention distributions thus provide a fine-grained map of how the model allocates importance across tokens. While often leveraged for interpretability~\cite{jain2019attention}, attention patterns can also reflect the model’s exposure to specific training examples. Tokens or phrases that appeared frequently during training may induce sharper and more consistent attention weights across layers, whereas unseen samples may produce more diffuse or unstable patterns.

\subsection{Attention as a Potential Privacy Signal}

Given its central role in shaping internal representations, attention can inadvertently encode information about the training data. Because attention determines how context is aggregated, repeated exposure to a sequence may cause the model to internalize a distinctive focus pattern for that input. These patterns—manifested through self-similarity, stability, or layer-wise consistency—can differ systematically between member and non-member examples.

This observation motivates our proposed framework, \textbf{AttenMIA}. Rather than treating attention solely as a visualization or interpretability tool, we regard it as a measurable signal of model familiarity. By quantifying the structure and perturbation sensitivity of attention maps, AttenMIA reveals subtle but consistent membership-dependent variations within the internal dynamics of transformer-based LLMs.

%% file: sections/threat_model.tex
\section{Threat Model}

We consider a standard membership inference setting in which an adversary aims to determine whether a target data sample was included in the training set of a deployed LLM. Our analysis focuses on transformer-based generative models.

\noindent \textbf{Adversary’s Goal:}  Given query access to a target model $f_\theta$, the adversary seeks to infer a binary membership variable $m(x) \in \{0,1\}$ for a data instance $x$. The attack is successful when the adversary can accurately distinguish between samples that were part of the training data of the model ($m(x)=1$) and those that were not ($m(x)=0$). The overall objective is to maximize inference accuracy under realistic assumptions of model access.

\noindent 
\textbf{Adversary’s Access and Knowledge:}
We assume a \emph{white-box} setting in which the adversary has full access to the internal components of the model, including attention-related state.  This assumption corresponds to a scenario where the model parameters are released (e.g., via open-weight LLMs like Llama2).  We do not assume that the attacker has access to the original training data or to the optimizer state.   %It can also apply to membership inference attacks on fine-tuned public models (e.g.. those available on repositories such as HuggingFace
Formally, the adversary can extract the attention weight matrices from target large language model, $f_\theta$, given an input sequence $x = (x_1, \ldots, x_T)$.  Specifically, these attention weight matrices are denoted $\{A^{(l,h)}\}_{l=1,h=1}^{L,H}$, where $A^{(l,h)} \in \mathbb{R}^{T \times T}$ represents the normalized attention map for head $h$ in layer $l$.
%\end{itemize}

\noindent This setting allows the adversary to probe how the model distributes its internal attention and contextual dependencies when processing a given input, thereby uncovering subtle indicators of training membership. Our threat model does not assume access to shadow models, distinguishing it from traditional reference-based MIA frameworks. Shadow models are auxiliary models trained by the adversary to imitate the target model’s behavior on synthetic or auxiliary data.  These models allow the attacker to learn how training membership affects model responses. While effective for small-scale classifiers, this approach is difficult for large language models, where training is computationally expensive and the underlying data distribution is unknown.  The attack is thus \emph{reference-free}: it relies solely on the intrinsic characteristics of the target model’s attention behavior. We also assume that no defensive mechanisms such as differential privacy, output perturbation, or attention masking have been applied.

%% file: sections/methodology.tex
\section{\textsc{AttenMIA} Overview and Design}

 \textbf{AttenMIA} is an attention-based approach for MIAs on transformer-based models such as large language models. The core insight is that self-attention patterns, thought to be a central mechanism for information flow, e.g., for facilitating contextual reasoning, also encode subtle but systematic signatures of memorization. By exploiting these signals, our approach is able to distinguish member from non-member samples with high accuracy.

% \subsection{Threat Model and Attacker Access}

% We assume a \textbf{white-box} adversary with access to the model’s attention tensors for any input sequence, but without access to output logits. Given a sequence $x = (t_1, t_2, \ldots, t_T)$ and transformer model $f$, the adversary can extract attention matrices $A^{\ell,h} \in \mathbb{R}^{T \times T}$ for each layer $\ell$ and head $h$. Each row $A^{\ell,h}_{i,:}$ is treated as an attention distribution over attended tokens. The adversary’s goal is to infer whether $x$ was in the training set.

\subsection{Intuition: Attention as a Membership Signal}

Our feature design is motivated by two hypotheses:\\ \noindent\textbf{H1.} Self-attention transitions across layers and heads encode membership signals: training samples should induce more consistent transition patterns, while non-members exhibit noisier and less structured flows.\\
\noindent \textbf{H2.} Members are expected to respond differently to \textit{perturbations}, since perturbations transform them from members to non-members, resulting in larger shifts away from the memorization attention patterns.  
%We explore these intuitions in this section.

\begin{figure}[t]
    \centering
    % --- Subfigure (a) ---
    \begin{subfigure}{\linewidth}
        \centering
        \includegraphics[width=\linewidth]{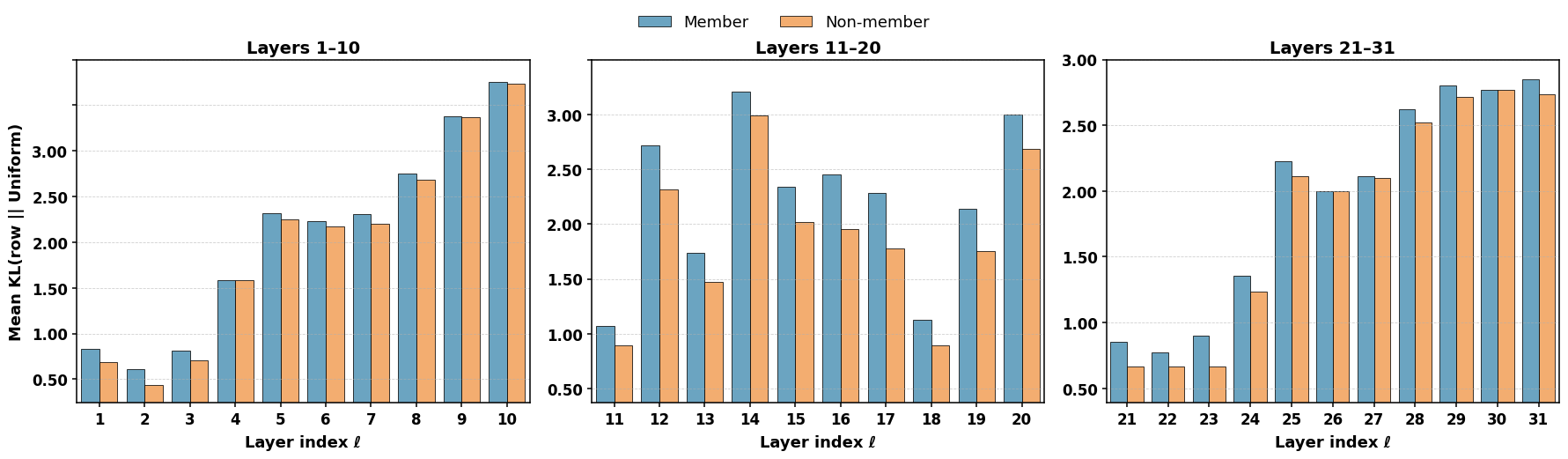}
        \caption{Per-layer attention activation (KL$\rightarrow$Uniform) for members vs.\ non-members in OPT-6.7B. Members exhibit consistently higher concentration in later layers.}%\nael{Absolute difference is really small; is there a way to show that this is significant?} \pedram{I showed the figure (same results) in a way that shows the difference more than the previous version.}
        \label{fig:visual_a}
    \end{subfigure}

    \vspace{1em} % Optional space between the two subfigures

    % --- Subfigure (b) ---
    \begin{subfigure}{\linewidth}
        \centering
        \includegraphics[width=\linewidth]{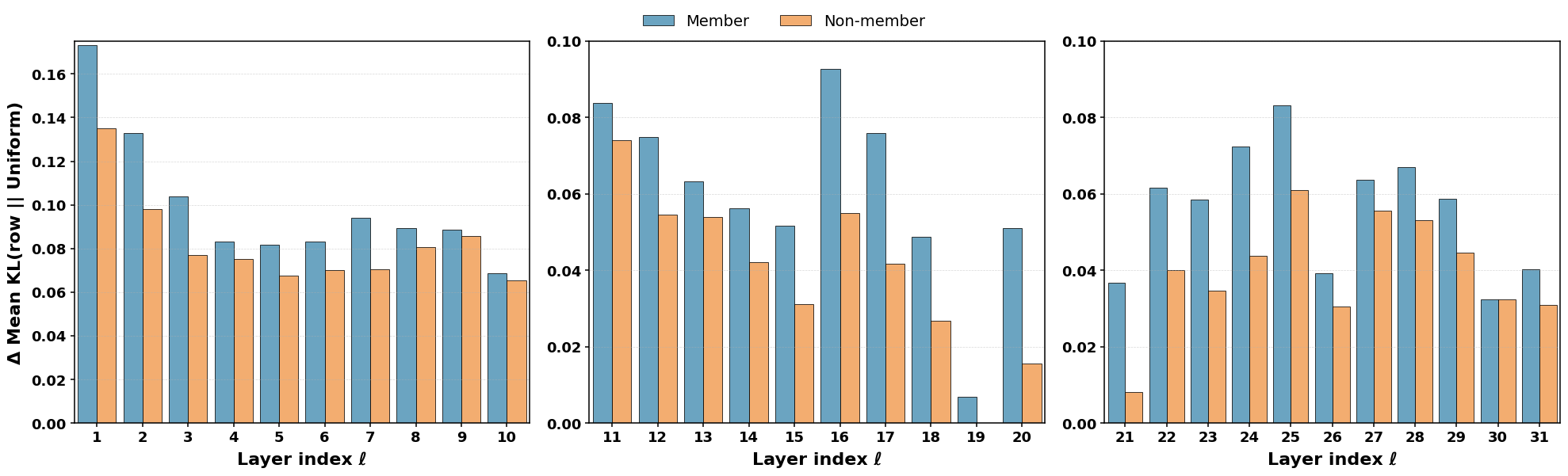}
        \caption{Per-layer relative change ($\Delta$ KL) under token-drop perturbations. Members undergo consistently larger shifts compared to non-members.}
        \label{fig:visual_b}
    \end{subfigure}

    \caption{Comparison of per-layer attention activation and perturbation effects in OPT-6.7B.}
    \label{fig:visual}
\end{figure}

We begin by examining average attention statistics on the WikiMIA benchmark. In Figure~\ref{fig:visual_a}, we compute the Kullback-Leibler (KL) divergence, which is an information-theoretic measure of relative entropy used to calculate the distance between two distributions.  Specifically, the KL divergence of each attention row is calculated against a uniform distribution, averaged across all heads in a layer, defined as:
\begin{equation}
    \kappa^{\ell,h} = \frac{1}{T} \sum_{i=1}^{T} \mathrm{KL}\!\left(A^{\ell,h}_{i,:} \,\|\, U_T\right),
\label{Eq: KL}
\end{equation}
where $A^{\ell,h}_{i,:}$ is the attention distribution of token $i$ in head $h$ of layer $\ell$, and $U_T$ is the uniform distribution over $T$ tokens. Larger values indicate sharper, less uniform attention. We observe that members consistently achieve higher divergence values than non-members, suggesting that the model allocates more concentrated attention to training samples, signaling overfitting/memorization. Moreover, while early layers attend nearly uniformly to both members and non-members, deeper layers exhibit increasingly focused distributions, with a clear separation between members and non-members. Although the magnitude of the difference is relatively small in some layers, these differences are consistent across all layers, and enable separation of member and non-member samples.  %\nael{These observations are a stretch from this figure.  I definitely wouldn't say that the separation is "clear".  How do we know that this is not just noise?  How about some phrasing such as: although the magnitude of the difference is small, these differences are consistent across all layers.} 

Next, we evaluate stability under perturbation. In Figure~\ref{fig:visual_b}, we drop seven tokens at fixed positions for both members and non-members, and measure the percentage change in their attention distributions layer by layer. Members generally exhibit larger shifts, reflecting that the model’s memorization makes their attention more sensitive to token removal. The earliest layers exhibit significant changes for both groups, whereas the deeper layers maintain a stronger separation between member and non-member responses.

\begin{figure*}[t]
    \centering
    \includegraphics[width=0.95\linewidth]{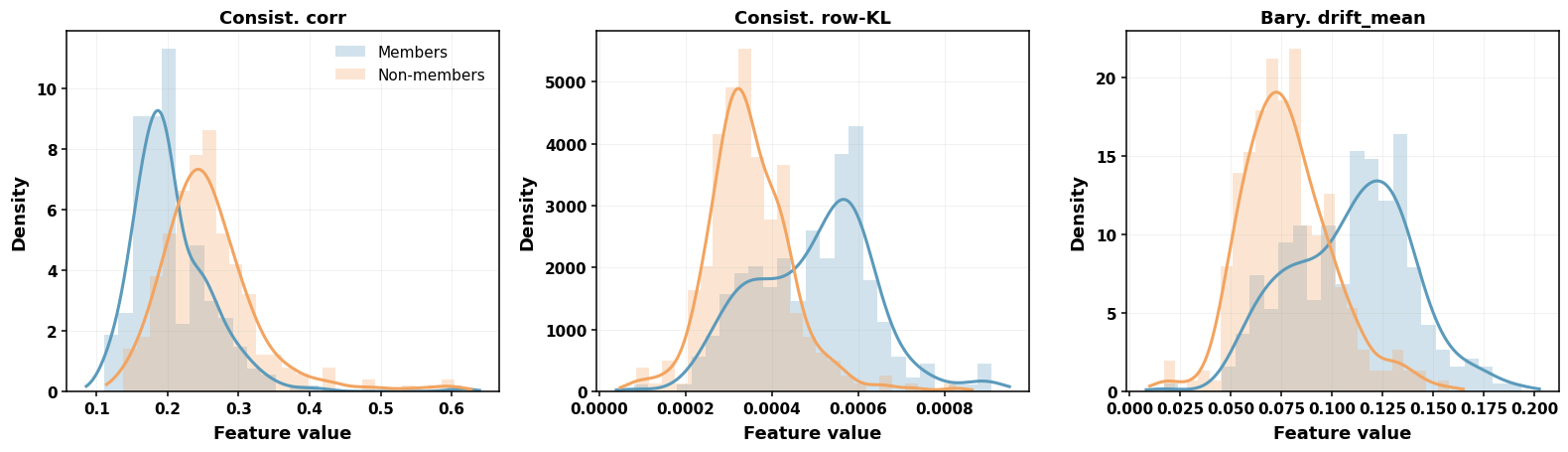}
    \caption{Kernel density estimates of transitional attention features for members (blue) and non-members (orange). We compute five metrics across adjacent heads and layers: correlation, Frobenius distance, row-wise KL divergence, and barycenter drift (mean and variance). In all cases, the distributions for members and non-members are clearly separable, indicating that transition dynamics encode reliable membership signals.}%\nael{Possible to show some traditional signal (e.g., output perplexity or whatever is used in LIRA) to show how these features compare to other traditional features?} \pedram{I have plotted for LIRA. I will add the figure.}
    \label{fig: Raw_features_KDE}
\end{figure*}

Finally, we study transitions between adjacent heads across layers. For each pair of consecutive heads, we compute five metrics that quantify distributional shifts, including correlation, Frobenius distance, row-wise KL divergence, and barycenter drift (mean and variance). Figure~\ref{fig: Raw_features_KDE} plots kernel density estimates of these features, showing systematic distributional shifts between members and non-members. In all cases, the two groups are separable, suggesting that transition features provide reliable discriminative power for training a membership classifier.  These statistics show substantially less overlap than output-based statistics used in prior MIA attacks.

To understand how well \textit{AttenMIA} features separate members from non-members, we evaluate these features on the GitHub subset of the MIMIR dataset using the Pythia-1.4B model, and compare them to the features used by the baselines. \emph{Loss} attack~\cite{yeom2018privacy} uses input loss as the membership score; \emph{Zlib}~\cite{carlini2021extracting} compresses the sequence of token losses using Zlib entropy; \emph{Neighbor}~\cite{mattern2023membership} compares each token’s loss to that of its nearest neighbors; LIRA~\cite{carlini2022membership} uses model likelihood variance across checkpoints to detect membership. Probability-based approaches include \emph{Min-K\%}~\cite{shi2023detecting}, which averages the minimum top-\(K\%\) predicted token probabilities across the input, and \emph{Min-K\%++}~\cite{zhang2024min}, which extends Min-K\% with normalization factors.

\noindent In this experiment, \textit{AttenMIA} uses only the transitional Consistency--Corr, Consistency--Frob, and Consistency--KL features computed between adjacent attention heads. The Hellinger Distance (HD)~\cite{nikulin2001hellinger} measures how different two distributions are, with a value of 0 indicating identical distributions (complete overlap), and a value of 1 indicating fully separated distributions. Table~\ref{tab:mia_hellinger_results} represents the Hellinger Distance for baselines and AttenMIA. AttenMIA illustrates that many individual heads show clear differences between members and non-members, with several achieving HD values between 0.60 and 0.70, exceeding other baselines' HD of $0.506$-$0.557$ on the same model/dataset. This superior separation between members and non-members supports the intuition that attention signals carry meaningful information about memorization and can outperform classical MIA methods.  Also contributing to \textsc{AttenMIA}'s advantage is the fact that we use a number of statistical features as well as input perturbations, and take advantage of information encoded in different heads and layers: the \emph{focus of attention within layers}, its \emph{stability under perturbations}, and its \emph{transition dynamics across heads and layers} all carry consistent signatures of membership. These findings motivate the selection of feature families, which we use as the basis of our MIA attacks in the next subsection.

% To understand how well \textit{AttenMIA} features separate members from non-members, we evaluate these features on the GitHub subset of the MIMIR dataset using the Pythia-1.4B model, and compare them to the features used by LIRA~\cite{carlini2022membership} as a baseline. In this experiment, \textit{AttenMIA} uses only the transitional Consistency--KL feature computed between adjacent attention heads. The Hellinger Distance (HD)~\cite{nikulin2001hellinger} measures how different two distributions are, with a value of 0 indicating identical distributions (complete overlap), and a value of 1 indicating fully separated distributions. Using Hellinger Distance, we find that many individual heads show clear differences between members and non-members, with several achieving HD values between 0.60 and 0.70, exceeding LIRA’s HD of $0.56$ on the same model/dataset. This superior separation between members and non-members supports the intuition that attention signals carry meaningful information about memorization and can outperform classical MIA methods.  Also contributing to \textsc{AttenMIA}'s advantage is the fact that we use a number of statistical features as well as input perturbations, and take advantage of information encoded in different heads and layers: the \emph{focus of attention within layers}, its \emph{stability under perturbations}, and its \emph{transition dynamics across heads and layers} all carry consistent signatures of membership. These findings motivate the selection of feature families, which we use as the basis of our MIA attacks in the next subsection.

\begin{table}[h!]
\centering
\resizebox{\linewidth}{!}{
\begin{tabular}{l c}
\toprule
\textbf{Method} & \textbf{Hellinger Distance} \\
\midrule

LIRA                & 0.557 \\
LOSS                & 0.532 \\
Zlib                & 0.557 \\
Neighbor            & 0.545 \\
MIN-K\%             & 0.532 \\
MIN-K\%++           & 0.506 \\
% RECALL              & \textemdash \\

\midrule

AttenMIA (Consistency–Corr)  & \textbf{0.704} \\
AttenMIA (Consistency–Frob)  & 0.663 \\
AttenMIA (Consistency–KL)    & 0.701 \\

\bottomrule
\end{tabular}
}
\caption{\textbf{Comparison of Membership Inference Attack Methods.}
We report the Hellinger Distance between member and non-member score distributions 
(higher is better). The AttenMIA variants achieve the highest HD values among 
all methods. Unlike baselines that compute a single score per sample, AttenMIA extracts many head features across the transformer, resulting in a richer signal and stronger separability. We also report the highest HD between the transformer heads.}
\label{tab:mia_hellinger_results}
\end{table}

% \begin{figure*}[t]
%     \centering
%     \includegraphics[width=0.95\linewidth]{img/visual.pdf}
%     \caption{a) Per-layer attention activation (KL$\rightarrow$Uniform) for members vs.\ non-members in OPT-6.7B. Members exhibit consistently higher concentration in later layers. b) Per-layer relative change ($\Delta$ KL) under token-drop perturbations. Members undergo consistently larger shifts compared to non-members.}
%     \label{fig: visual}
% \end{figure*}

% \subsection{Raw Attention Features}

% Let $A^{\ell,h} \in \mathbb{R}^{T \times T}$ denote the attention weights of layer $\ell$ and head $h$. We extract the following statistics across adjacent layers $(\ell, \ell+1)$:

% \paragraph{Consistency.} Measures stability of attention patterns between layers:
% \begin{align}
% \mathrm{Corr}^{\ell,h} &= \mathrm{corr}\!\left(\mathrm{vec}\,A^{\ell,h}, \,\mathrm{vec}\,A^{\ell+1,h}\right), \\
% \Delta_F^{\ell,h} &= \frac{\lVert A^{\ell+1,h} - A^{\ell,h}\rVert_F}{T^2}, \\
% \Delta_{\mathrm{KL}}^{\ell,h} &= \frac{1}{T}\sum_{i=1}^{T} \mathrm{KL}\!\left(A^{\ell,h}_{i,:} \,\big\|\, A^{\ell+1,h}_{i,:}\right).
% \end{align}

% \paragraph{Barycenter Drift.} Captures shifts in the focus of attention across positions. For row $i$, define the barycenter:
% \[
% c_i^{\ell,h} = \sum_{j=1}^{T} j \cdot A^{\ell,h}_{i,j}.
% \]
% We compute the mean and variance of layer-to-layer drifts:
% \begin{align}
% d_i^{\ell,h} &= |c_i^{\ell+1,h} - c_i^{\ell,h}|, \qquad
% \overline{d}^{\ell,h} = \tfrac{1}{T}\sum_i d_i^{\ell,h}, \qquad
% \mathrm{Var}_d^{\ell,h} = \mathrm{Var}_i(d_i^{\ell,h}).
% \end{align}

\subsection{Transitional Attention Features}

To complement signals targeting attention distributions in each layer, we consider features tracking attention changes across layers.  Let $A^{\ell,h} \in \mathbb{R}^{T \times T}$ denote the attention matrix of layer $\ell$ and head $h$, where $T$ is the sequence length. We consider two families of statistics across adjacent layers $(\ell, \ell+1)$: \emph{consistency} and \emph{barycenter drift}. These features capture the stability and positional dynamics of attention transitions. \\

\noindent \textbf{Consistency}. This family measures how stable the attention distributions remain across consecutive layers. We define three complementary metrics:

\begin{align}
\textbf{Consistency–Corr:} \quad 
\mathrm{Corr}^{\ell,h} &= \mathrm{corr}\!\left(\mathrm{vec}\,A^{\ell,h}, \,\mathrm{vec}\,A^{\ell+1,h}\right),
\end{align}

\begin{align}
\textbf{Consistency–Frob:} \quad 
\Delta_F^{\ell,h} &= \frac{\lVert A^{\ell+1,h} - A^{\ell,h}\rVert_F}{T^2},
\end{align}

\begin{align}
\textbf{Consistency–KL:} \quad 
\Delta_{\mathrm{KL}}^{\ell,h} &= \frac{1}{T}\sum_{i=1}^{T} \mathrm{KL}\!\left(A^{\ell,h}_{i,:} \,\big\|\, A^{\ell+1,h}_{i,:}\right).
\end{align}

We selected these three metrics because describe different aspects of how attention patterns change between layers. Correlation reflects how similar the overall structure of two attention maps is, Frobenius distance measures the magnitude of their difference, and row-wise KL divergence shows how token-level attention distributions shift. Together, these metrics offer complementary views of layer-to-layer dynamics, giving a more complete picture of attention consistency than any single measure alone. %} \nael{Why pick these three?  Why do we need all three--do they provide complementary views?}\\

\noindent \textbf{Barycenter Drift.} To measure positional shifts in attention, we compute the barycenter of each row:
\[
c_i^{\ell,h} = \sum_{j=1}^{T} j \cdot A^{\ell,h}_{i,j}.
\]
The layer-to-layer drift for token $i$ is then
\[
d_i^{\ell,h} = |c_i^{\ell+1,h} - c_i^{\ell,h}|,
\]
and we summarize these drifts with two statistics:
\begin{align}
\textbf{Barycenter–Mean:} \quad 
\overline{d}^{\ell,h} &= \tfrac{1}{T}\sum_{i=1}^T d_i^{\ell,h}, \\
\textbf{Barycenter–Var:} \quad 
\mathrm{Var}_d^{\ell,h} &= \mathrm{Var}_i(d_i^{\ell,h}).
\end{align}
These features reflect how much the center of attention shifts on average (mean drift) and how uniformly it shifts across tokens (variance) across consecutive layers.

\subsection{Perturbation-Based Analysis}

Prior MIA attacks have explored an active perturbation-driven membership inference approach~\cite{xie2024recall} to distinguish members and non-members. The features considered so far, characterize attention with respect to a given input sample, capturing structural differences between members and non-members.   Input perturbations, where we change the input sample and observe changes in the attention pattern, provide another complementary source of attention-based signals. We modify this approach for our attack, applying the input perturbations but observing the change in the distribution of the attention signals (rather than the output logits).  The intuition is that training samples are more deeply embedded in the model parameters, and therefore, their attention distributions respond differently to perturbations compared to non-members. An alternative view is that perturbations more asymmetrically affect attention patterns for member samples as the perturbations move them from being members to non-members (while non-member samples remain non-members). 

%\nael{I recall there was another paper that used this approach.  We need to credit it if so.}

We consider three heuristics for perturbations, recognizing that others are possible:
\begin{itemize}
    \item \textbf{Token Dropping:} remove a fixed number of tokens at predetermined positions.
    \item \textbf{Token Replacement:} substitute selected tokens with unrelated vocabulary items.
    \item \textbf{Prefix Insertion:} prepend an unrelated non-member sample as a prefix.
\end{itemize}

\noindent For a given input $x$ and its perturbed counterpart $x'$, we define the KL concentration shift score $\Delta \kappa^{\ell,h}$ for each head as:

\begin{equation}
    \Delta \kappa^{\ell,h} = \frac{1}{T} \sum_{i=1}^{T} \mathrm{KL}\!\left(A^{\ell,h}_{i,:} \,\|\, A'^{\ell,h}_{i,:}\right),
\label{Eq: KL}
\end{equation}

\begin{figure*}[t]
    \centering
    \includegraphics[width=0.95\linewidth]{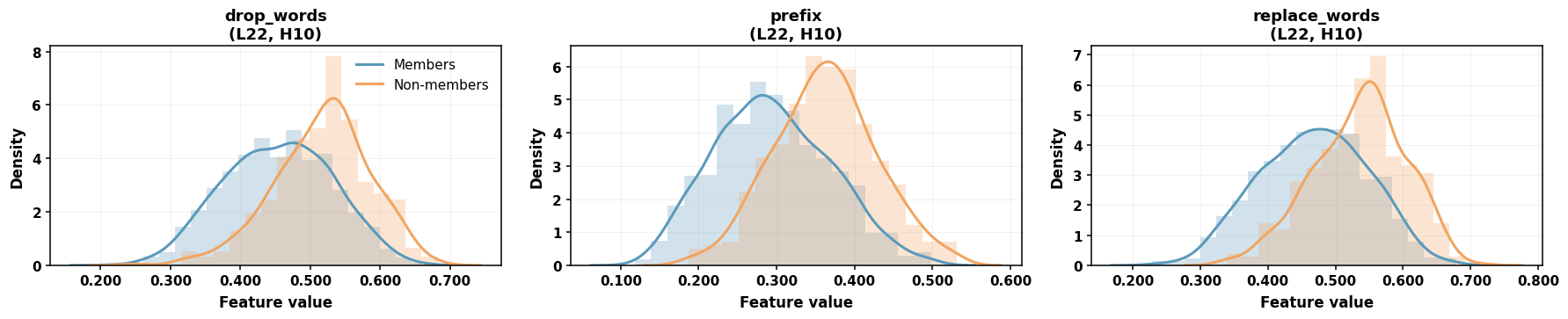}
    \caption{KDE plots of head-level correlation of attention distribution under perturbation for members (blue) and non-members (orange). Distributions are clearly separated, highlighting that correlation shifts provide strong signals for membership inference. Results are shown on the Arxiv subset of MIMIR with the Pythia-1.4B model.}
    \label{fig: perturb_kde}
\end{figure*}

\noindent Intuitively, members tend to maintain more stable attention concentration under perturbations, whereas non-members exhibit larger shifts. Figure~\ref{fig: perturb_kde} confirms this intuition: the kernel density estimates (KDEs) of correlation values for members and non-members show clear distributional separation across multiple layers and perturbation types.

\subsection{Feature Aggregation and Classifier}

%\nael{The classifier seems a bit ad hoc.  It's not clear why this approach that fuses everything is best.  Perhaps we should evaluate individual features, then their combination.  We should also think about identifying the most salient heads/layers and classifying based on that.  I see later that you have ablation studies to see the effects of different features and layers.  But we need a discussion here to explain why this is the design of the classifier.}
Given the set of attention features, including those from input perturbations, a natural question is how do we exploit these features to build a membership classifier that can identify potential samples that are likely to be part of the training data set.  We discuss our approach in this section. %While it is possible to use any subset of features, or even more granular features (e.g., focus on certain layers/heads), we elect to use an ensemble of the features presented so far.  Specifically, the final stage of \textsc{AttenMIA} combines all extracted features into a unified representation for classification. 

\textbf{Classifier design and Feature Selection}: Our classifier design was built based on empirical experimentation with different attention features and different perturbation strategies.  First, evaluating each perturbation strategy independently shows that, although each contributes a meaningful membership signal, their combination consistently enhances both ROC-AUC and TPR@1\%FPR, demonstrating that the perturbations capture complementary aspects of model behavior. 
Second, layer-wise analysis indicates that, in general, different layers encode useful information, and aggregating representations across layer depths yields higher accuracy than relying on individual layers or subsets of layers.   These evaluations imply that membership cues are distributed across perturbation types and model depths, and combining them provides the most reliable discrimination at low false-positive rates.

For each input sample $x$, we concatenate each of the following feature sets:
\begin{itemize}
    \item \textbf{Transitional attention features:} consistency (correlation, Frobenius distance, KL divergence), and barycenter drift (mean, variance).
    \item \textbf{Perturbation-based features:} KL-shift values $\Delta \kappa^{\ell,h}$ across the three perturbation strategies (token dropping, token replacement, prefix insertion).
\end{itemize}

\noindent This approach yields a high-dimensionality feature vector $\mathbf{z}$. To classify samples, we employ a lightweight multi-layer perceptron (MLP), trained using n-fold cross-validation to ensure balanced evaluation across members and non-members.  
We report standard metrics: (i) ROC AUC, measuring global separability, and (ii) TPR@1\%FPR (True positive rate at 1\% false positive rate), which highlights performance in the low-false-positive regime crucial for privacy auditing. Algorithm~\ref{alg: attenmia} represents our MIA algorithm.

\begin{algorithm}[t]
\caption{\textsc{AttenMIA}: Attention-Based Membership Inference}
\label{alg: attenmia}
\KwIn{Target LLM $\mathcal{M}$, training set with Members $\mathcal{D}_{\text{mem}}$ and Non-Members $\mathcal{D}_{\text{non}}$, perturbation strategies $\mathcal{P} = \{\text{prefix}, \text{drop}, \text{replace}\}$.}
\KwOut{Trained MIA classifier $\mathcal{C}$.}

\BlankLine
\textbf{Step 1: Feature Extraction}\\
\ForEach{sample $x \in \mathcal{D}_{\text{mem}} \cup \mathcal{D}_{\text{non}}$}{
    \tcp{Transitional attention features}
    Compute $F_{\text{trans}}(x)$ from $\mathcal{M}$\;
    
    \tcp{Perturbation-based features}
    \ForEach{perturbation $p \in \mathcal{P}$}{
        Generate perturbed input $x_p \leftarrow p(x)$\;
        Extract features $F_{\text{pert}}(x, p)$ from $\mathcal{M}$\;
    }
}

\BlankLine
\textbf{Step 2: Training the Classifier}\\
Construct training data:
\begin{itemize}
    \item Transitional feature vectors $F_{\text{trans}}(x)$
    \item Perturbation-based features $\{F_{\text{pert}}(x)\}_{p \in \mathcal{P}}$
\end{itemize}

Train MIA classifier $\mathcal{C}$ using labels 
(Member = 1, Non-Member = 0)\;

\BlankLine
\textbf{Step 3: Inference}\\
Given a test sample $x$:
\begin{enumerate}
    \item Extract features: $F_{\text{trans}}(x)$ or $F_{\text{pert}}(x)$ for $p \in \mathcal{P}$\;
    \item Predict membership: $\hat{y} \leftarrow \mathcal{C}(features)$\;
\end{enumerate}

\end{algorithm}

%% file: sections/experiments.tex
\section{Experimental Evaluation}

We evaluate AttenMIA on several diverse transformer architectures and datasets. Our experiments are designed to answer three questions:
(1) Do attention-based features provide reliable membership signals across models and datasets?
(2) How does AttenMIA compare with state-of-the-art MIAs?
(3) Which attention features and perturbation strategies contribute most to MIA performance?

\begin{table*}[t]
\centering
\small
\resizebox{\textwidth}{!}{%
\begin{tabular}{c|lcccccccccccc}
\toprule
Length & Method 
& \multicolumn{2}{c|}{Pythia-6.9B} 
& \multicolumn{2}{c|}{LLaMA-13B} 
& \multicolumn{2}{c|}{NeoX-20B} 
& \multicolumn{2}{c|}{LLaMA-30B} 
& \multicolumn{2}{c|}{OPT-66B} 
& \multicolumn{2}{c}{Average} \\
\cmidrule(lr){3-4} \cmidrule(lr){5-6} \cmidrule(lr){7-8} 
\cmidrule(lr){9-10} \cmidrule(lr){11-12} \cmidrule(lr){13-14} 

& & AUC & TPR & AUC & TPR & AUC & TPR & AUC & TPR & AUC & TPR & AUC & TPR  \\
\midrule

\multirow{10}{*}{32}
& PPL & 64.1 & 6.2 & 68.2 & 1.3 & 71.0 & 2.1 & 71.1 & 4.9 & 68.3 & 3.4 & 68.5 & 3.6 \\
& Loss         & 63.6 & 6.1 & 67.6 & 4.8 & 68.7 & 10.4 & 69.5 & 4.3 & 65.4 & 6.4 & 65.9 & 6.1 \\
& Ref         & 63.7 & 6.9 & 67.7 & 5.9 & 68.9 & 10.1 & 70.0 & 2.7 & 65.8 & 6.7 & 66.2 & 6.1 \\
& Zlib         & 64.1 & 4.8 & 67.8 & 5.6 & 68.9 & 9.1  & 69.9 & 4.8 & 65.5 & 5.6 & 66.3 & 5.7 \\
& Neighbor     & 65.8 & 0.8   & 65.8 & 0.6  & 70.2 & 0.6 & 67.6 & 0.6  & 68.2 & 0.8   & 66.9 & 0.7   \\
& Min-K\%      & 66.3 & 8.8 & 68.0 & 5.1 & 71.8 & 10.7 & 70.1 & 4.5 & 67.4 & 9.1 & 67.8 & 7.5 \\
& Min-K\%++   & 70.3 & 5.9 & 84.8 & 10.4 & 75.1 & 6.1 & 84.3 & 9.3 & 70.3 & 3.7 & 75.3 & 6.6 \\
& PETAL & 64.1 & 4.9 & 58.0 & 1.6 & 62.1 & 2.1 & 60.0 & 1.9 & 62.1 & 3.1 & 61.3 & 2.7 \\
& RECALL  & 91.6 & 28.5 & 92.2 & 13.3 & 90.5 & 25.3 & 90.7 & 18.4 & 85.1 & 8.3 & 90.1 & 17.5 \\

& AttenMIA (Transitional)     & \textbf{98.9} & \textbf{86.4} & 99.3 & 84.8 & \textbf{99.4}  & \textbf{86.8} & 99.7 & 94.1 & 97.6 & 84.2 & \textbf{99.0} & \textbf{87.3} \\

& AttenMIA (Perturbed)  & 98.2 & 66.0 & \textbf{99.6} & \textbf{87.9} & 98.9 & 86.4 & \textbf{99.7} & \textbf{94.7} & \textbf{97.8} & \textbf{84.6} & 98.8 & 83.9 \\

\midrule

\multirow{10}{*}{64}
& PPL & 62.2 & 4.8 & 64.0 & 2.1 & 68.0 & 1.3 & 67.1 & 3.6 & 64.0 & 3.2 & 65.1 & 3.0 \\
& Loss         & 61.7 & 3.3 & 64.4 & 4.9 & 67.4 & 4.5 & 66.7 & 6.1 & 63.1 & 4.1 & 63.8 & 4.4 \\
& Ref         & 61.8 & 3.3 & 64.9 & 4.1 & 67.7 & 4.9 & 67.6 & 6.5 & 63.6 & 4.5 & 64.2 & 4.4 \\
& Zlib        & 63.3 & 6.9 & 65.9 & 8.9 & 68.7 & 7.7 & 68.0 & 10.6 & 64.6 & 9.8 & 65.3 & 8.3 \\
& Neighbor    & 63.2 & 0.6 & 64.1 & 0.6 & 67.1 & 0.8 & 67.1 & 0.6 & 64.1 & 0.6 & 64.4 & 0.6   \\
& Min-K\%   & 65.0 & 6.5 & 66.9 & 6.5 & 73.5 & 5.7 & 69.1 & 8.1 & 67.9 & 10.2 & 67.6 & 7.3 \\
& Min-K\%++   & 71.7 & 11.8 & 85.6 & 15.4 & 76.0 & 10.2 & 84.8 & 6.9 & 70.2 & 11.8 & 75.9 & 10.6 \\
& PETAL & 63.0 & 4.2 & 56.2 & 1.4 & 62.0 & 2.1 & 58.3 & 1.8 & 62.0 & 2.9 & 60.3 & 2.5 \\
& RECALL   & 93.0 & 20.7 & 95.2 & 30.1 & 93.2 & 6.9 & 94.9 & 18.3 & 79.9 & 5.3 & 91.3 & 15.4 \\
& AttenMIA (Transitional)        & \textbf{97.6} & 58.5 & 97.9 & 67.5 & \textbf{98.4} & \textbf{74.9} & \textbf{99.1} & 75.6 & 96.4 & \textbf{74.2} & \textbf{97.9} & 70.1 \\

& AttenMIA (Perturbed)   & 96.6 & \textbf{60.1} & \textbf{98.5} & \textbf{74.5} & 97.8 & 65.2 & 99.0 & \textbf{80.2} & \textbf{96.8} & 73.1 & 97.7 & \textbf{70.6} \\

\midrule

\multirow{10}{*}{128} 
& PPL & 61.0 & 3.8 & 61.8 & 1.3 & 64.0 & 1.3 & 65.1 & 3.2 & 61.0 & 2.9 & 62.6 & 2.5 \\
& Loss         & 65.0 & 3.0 & 69.1 & 7.1 & 70.6 & 4.0 & 72.0 & 1.0 & 65.3 & 4.0 & 67.5 & 3.4 \\
& Ref         & 65.1 & 3.0 & 69.3 & 8.1 & 70.8 & 4.0 & 73.0 & 0.0 & 65.5 & 4.0 & 67.8 & 3.4 \\
& Zlib        & 67.8 & 6.1 & 71.5 & 10.1 & 72.6 & 5.1 & 73.6 & 2.0 & 67.6 & 9.1 & 69.8 & 6.4 \\
& Neighbor     & 67.5 & 0.4 & 68.3 & 0.8  & 71.6 & 0.6 & 72.2 & 0.6 & 67.7 & 0.8 & 68.7 & 0.6   \\
& Min-K\%     & 69.5 & 4.0 & 71.5 & 8.1 & 75.0 & 3.0 & 74.2 & 2.0 & 70.2 & 4.0 & 71.2 & 4.0 \\
& Min-K\%++    & 69.7 & 8.1 & 83.9 & 8.1 & 75.8 & 1.0 & 82.9 & 0.0 & 72.1 & 0.0 & 75.2 & 3.2 \\
& PETAL & 61.1 & 3.7 & 56.0 & 1.3 & 60.0 & 1.8 & 56.1 & 1.6 & 60.2 & 2.6 & 58.7 & 2.2 \\
& RECALL  & 92.6 & 33.3 & 92.5 & 26.3 & 91.7 & 30.3 & 91.2 & 1.0 & 81.0 & 6.1 & 90.0 & 16.9 \\

& AttenMIA (Transitional)        & \textbf{92.8} & \textbf{39.8} & 94.8  & 39.9  & \textbf{95.3} & \textbf{53.2} & 96.8 & 57.3 & 94.2 & \textbf{54.4} & \textbf{94.8} & \textbf{48.9} \\

& AttenMIA (Perturbed)   & 91.5 & 17.5 & \textbf{97.1} & \textbf{61.0} & 86.6 & 36.5 & \textbf{97.5} & \textbf{58.3} & \textbf{94.4} & 53.8 & 93.4 & 45.4 \\

\bottomrule

\end{tabular}}
\caption{ROC AUC and TPR@1\%FPR on the WikiMIA benchmark across sequence lengths (32, 64, 128 tokens) and five LLMs. Each model column reports AUC and TPR@1\%FPR. AttenMIA (Transitional and Perturbed) consistently achieves the best performance across all models and lengths, with large gains in the low-FPR regime. Results are averaged across evaluation folds.}
\label{tab:main_results_wikimia}
\end{table*}

\subsection{Experimental Setup}

\noindent \textbf{Datasets and Models.} 
We evaluated AttenMIA on two datasets commonly used for membership inference research: WikiMIA \cite{shi2023detecting} and MIMIR \cite{duan2024membership}. Both datasets provide gold membership labels; that is, labels that identify the ground truth on whether each sample is a member of the training dataset.  For each dataset, we perform 5-fold cross-validation: the data is partitioned into five folds, and in each run, one fold is used for testing while the others are used for training the classifier model. We report mean performance across folds to reduce variance and ensure robustness.  We evaluate attacks on several open-weight transformer models, including Pythia \cite{biderman2023pythia}, GPT-NeoX \cite{black2022gpt}, OPT \cite{zhang2022opt}, and LLaMA-2 \cite{touvron2023llama}, spanning sizes from 1.4B to 66B parameters. This range allows us to study both architectural and scaling effects. Unless noted otherwise, all models are used in their released form without fine-tuning.

\noindent \textbf{Evaluation Metrics.} 
Following prior works, we report two standard metrics for membership inference: (i) ROC AUC, which measures overall discriminative ability; and (ii) TPR@1\%FPR, which captures performance in the low-false-positive regime crucial for privacy auditing. A high TPR\@1\%FPR indicates that training samples can be identified with high confidence while triggering few false alarms. All metrics are averaged across the cross-validation folds. \\

\noindent \textbf{Baselines.} 
We compare \textsc{AttenMIA} against several strong membership inference attacks. RECALL~\cite{xie2024recall} introduced input perturbations, comparing relative conditional likelihood shifts induced by non-member prefixes, where member samples exhibit larger likelihood drops. PETAL~\cite{mattern2023membership} leverages token-level semantic similarity to approximate output probabilities and subsequently calculate the perplexity. It exposes membership based on the common assumption that members are better memorized and have smaller perplexity. \emph{Loss} attack~\cite{yeom2018privacy} uses input loss as the membership score, and \emph{PPL} attack~\cite{carlini2021extracting} uses output probabilities (perplexity) directly under the assumption that memorized samples yield lower perplexity. \emph{Zlib}~\cite{carlini2021extracting} compresses the sequence of token losses using Zlib entropy; \emph{Ref}~\cite{carlini2022membership} calibrates the input loss of the target model using a reference model; \emph{Neighbor}~\cite{mattern2023membership} compares each token’s loss to that of its nearest neighbors. 
Probability-based approaches include \emph{Min-K\%}~\cite{shi2023detecting}, which averages the minimum top-\(K\%\) predicted token probabilities across the input, and \emph{Min-K\%++}~\cite{zhang2024min}, which extends Min-K\% with normalization factors.

\subsection{Main Results}
\label{sec:main_results}

\noindent \textbf{Overall performance in AUC and TPR@1\%FPR.} 
Table~\ref{tab:main_results_wikimia} summarizes the performance of \textsc{AttenMIA} and existing baselines on WikiMIA, while Tables~\ref{tab:main_results_mimir_AUC} present complementary results across Pythia variants and the MIMIR benchmark (more results in Appendix~\ref{sec: big_table}, Table~\ref{tab:main_results_mimir_TPR} ).

Across all models, datasets, and evaluation settings, \textsc{AttenMIA} consistently outperforms prior membership inference attacks. On the MIMIR, averaged across datasets and model scales, \textsc{AttenMIA} improves ROC AUC by \textbf{+55\%} over RECALL and \textbf{+27\%} over PETAL (Table~\ref{tab:main_results_mimir_AUC}). In the low-FPR regime, critical for privacy auditing, \textsc{AttenMIA} achieves an average TPR@1\%FPR of \textbf{47.8\%}, surpassing both RECALL (19.6\%) and PETAL (13.3\%) (Table~\ref{tab:main_results_mimir_TPR}). 
These gains highlight that attention-based features capture memorization signals that are not visible from surface-level outputs, enabling more accurate and reliable auditing across diverse LLM families.

% ------------------------------------------------------------------------------
% ------------------------------------------------------------------------------

% \renewcommand{\arraystretch}{1.5}
\begin{table*}[t]
\centering
\resizebox{\textwidth}{!}{
\begin{tabular}{lcccccccccccc}
\multicolumn{13}{c}{} \\
\toprule
\multirow{2}{*}{\textbf{Method}} &
\multicolumn{3}{c}{\textbf{Wikipedia (Pythia)}} &
\multicolumn{3}{c}{\textbf{GitHub (Pythia)}} &
\multicolumn{3}{c}{\textbf{Pile CC (Pythia)}} &
\multicolumn{3}{c}{\textbf{PubMed (Pythia)}} \\
\cmidrule(lr){2-4} \cmidrule(lr){5-7} \cmidrule(lr){8-10} \cmidrule(lr){11-13}

& 1.4b & 2.8b & 6.9b & 1.4b & 2.8b & 6.9b & 1.4b & 2.8b & 6.9b & 1.4b & 2.8b & 6.9b \\

\midrule

PPL attack & 0.62 & 0.63 & 0.64 & 0.86 & 0.87 & 0.88 & 0.52 & 0.53 & 0.54 & 0.69 & 0.68 & 0.70  \\

reference attack & 0.58 & 0.59 & 0.59 & 0.65 & 0.65 & 0.64 & 0.51 & 0.51 & 0.53 & 0.65 & 0.64 & 0.65  \\

zlib attack  & 0.60 & 0.61 & 0.62 & 0.85 & 0.86 & 0.87 & 0.51 & 0.52 & 0.52 & 0.71 & 0.70 & 0.72  \\

neighborhood attack & 0.56 & 0.56 & 0.57 & 0.83 & 0.83 & 0.84 & 0.50 & 0.50 & 0.51 & 0.69 & 0.68 & 0.69  \\

MIN-K\% PROB  & 0.59 & 0.59 & 0.62 & 0.86 & 0.86 & 0.88 & 0.52 & 0.52 & 0.52 & 0.53 & 0.53 & 0.55  \\

WS attack  & 0.53 & 0.55 & 0.52 & 0.76 & 0.76 & 0.77 & 0.50 & 0.49 & 0.50 & 0.52 & 0.54 & 0.54  \\

RS attack  & 0.54 & 0.54 & 0.55 & 0.77 & 0.77 & 0.78 & 0.50 & 0.48 & 0.49 & 0.54 & 0.54 & 0.56    \\

BT attack  & 0.53 & 0.54 & 0.52 & 0.74 & 0.76 & 0.77 & 0.52 & 0.50 & 0.50 & 0.52 & 0.53 & 0.56 \\

PETAL  & 0.62 & 0.61 & 0.61 & 0.85 & 0.85 & 0.87 & 0.54 & 0.55 & 0.54 & 0.65 & 0.65 & 0.66  \\

RECALL & 0.52 & 0.52 & 0.54 & 0.70 & 0.72 & 0.74 & 0.52 & 0.50 & 0.52 & 0.52 & 0.50 & 0.51 \\

\midrule

AttenMIA (Transitional)  & 0.63 & 0.65 & 0.61 & 0.88 & 0.93 & 0.89 & 0.59 & 0.58 & 0.56 & 0.85 & 0.85 & 0.84  \\

AttenMIA (Perturbed)  & \textbf{0.74} & \textbf{0.73} & \textbf{0.73} & \textbf{0.99} & \textbf{0.99} & \textbf{1.0} & \textbf{0.66} & \textbf{0.70} & \textbf{0.71} & \textbf{0.95} & \textbf{0.96} & \textbf{0.97}  \\

\midrule
\multicolumn{13}{c}{} \\
\midrule
\multirow{2}{*}{\textbf{Method}} &
\multicolumn{3}{c}{\textbf{arXiv (Pythia)}} &
\multicolumn{3}{c}{\textbf{DM Math (Pythia)}} &
\multicolumn{3}{c}{\textbf{HackerNews (Pythia)}} &
\multicolumn{3}{c}{\textbf{Average (Pythia)}} \\
\cmidrule(lr){2-4} \cmidrule(lr){5-7} \cmidrule(lr){8-10} \cmidrule(lr){11-13}
& 1.4b & 2.8b & 6.9b & 1.4b & 2.8b & 6.9b & 1.4b & 2.8b & 6.9b & 1.4b & 2.8b & 6.9b \\
\midrule

PPL attack  & 0.66 & 0.64 & 0.66 & 0.88 & 0.87 & 0.86 & 0.59 & 0.60 & 0.6- & 0.69 & 0.69 & 0.70  \\

reference attack  & 0.67 & 0.64 & 0.67 & 0.82 & 0.78 & 0.77 & 0.52 & 0.53 & 0.52 & 0.63 & 0.62 & 0.62 \\

zlib attack  & 0.65 & 0.64 & 0.66 & 0.80 & 0.78 & 0.77 & 0.59 & 0.60 & 0.59 & 0.68 & 0.68 & 0.68 \\

neighborhood attack  & 0.69 & 0.69 & 0.69 & 0.73 & 0.72 & 0.73 & 0.53 & 0.55 & 0.54 & 0.65 & 0.65 & 0.65 \\

MIN-K\% PROB   & 0.62 & 0.61 & 0.62 & 0.71 & 0.67 & 0.68 & 0.57 & 0.58 & 0.59 & 0.63 & 0.62 & 0.64 \\

WS attack & 0.50 & 0.50 & 0.50 & 0.45 & 0.49 & 0.44 & 0.51 & 0.51 & 0.48 & 0.54 & 0.55 & 0.54 \\

RS attack  & 0.50 & 0.48 & 0.51 & 0.42 & 0.45 & 0.43 & 0.49 & 0.51 & 0.51 & 0.54 & 0.54 & 0.55 \\

BT attack  & 0.51 & 0.51 & 0.55 & 0.39 & 0.45 & 0.39 & 0.51 & 0.52 & 0.53 & 0.53 & 0.54 & 0.55 \\

PETAL  & 0.58 & 0.57 & 0.58 & 0.87 & 0.85 & 0.85 & 0.59 & 0.58 & 0.58 & 0.67 & 0.67 & 0.67  \\

RECALL  & 0.53 & 0.53 & 0.55 & 0.53 & 0.51 & 0.51 & 0.53 & 0.53 & 0.54 & 0.55 & 0.54 & 0.56  \\

\midrule

AttenMIA (Transitional)   & 0.81 & 0.79 & 0.78 & 0.94 & 0.95 & 0.96 & 0.62 & 0.59 & 0.59 & 0.76 & 0.76 & 0.75  \\

AttenMIA (Perturbed)  & \textbf{0.93} & \textbf{0.94} & \textbf{0.94} & \textbf{1.0} & \textbf{1.0} & \textbf{1.0} & \textbf{0.70} & \textbf{0.72} & \textbf{0.72} & \textbf{0.83} & \textbf{0.86} & \textbf{0.87}  \\

\bottomrule
\end{tabular}
}
\caption{ROC AUC of membership inference attacks across seven datasets and three Pythia model sizes (1.4B, 2.8B, 6.9B). AttenMIA (Transitional and Perturbed) consistently achieves the highest performance across all settings, outperforming baselines. Bold values indicate the best result for each dataset–model configuration.}
\label{tab:main_results_mimir_AUC}
\end{table*}

% ------------------------------------------------------------------------------
% -----------------------------------------------------------------------------

\noindent \textbf{Impact of input perturbation strategies.} 
We next study how different perturbation strategies influence attack performance. 
We evaluate three perturbations: (i) non-member prefix insertion, (ii) token dropping, and (iii) token replacement. These perturbations simulate natural variations in phrasing and test the stability of attention-based membership signals.
As shown in Table~\ref{tab:ablation_results_perturbations}, all three perturbation types yield comparable performance individually.  Moreover, combining the strategies produces incremental improvements leading to the strongest results.

\begin{table*}[h]
\centering
\resizebox{\textwidth}{!}{
\begin{tabular}{lcccccccccccc}
\multicolumn{9}{c}{} \\
\toprule
\multirow{2}{*}{\textbf{Method}} &
\multicolumn{2}{c}{\textbf{Wikipedia}} &
\multicolumn{2}{c}{\textbf{GitHub}} &
\multicolumn{2}{c}{\textbf{Pile CC}} &
\multicolumn{2}{c}{\textbf{PubMed Central}} \\
\cmidrule(lr){2-3} \cmidrule(lr){4-5} \cmidrule(lr){6-7} \cmidrule(lr){8-9}
& ROC AUC & TPR@1\%FPR & ROC AUC & TPR@1\%FPR & ROC AUC & TPR@1\%FPR & ROC AUC & TPR@1\%FPR \\
\midrule

AttenMIA (Non-Member Prefix)  & 0.61 & 11.5 & 0.92 & 73.6 & 0.60 & 6.5 & 0.86 & 24.5  \\
AttenMIA (Dropping Tokens)  & 0.62 & 11.5 & 0.92 & 74.3 & 0.59 & 5.1 & 0.86 & 27.5  \\
AttenMIA (Replacing Tokens)  & 0.62 & 11.6 & 0.92 & 72.2 & 0.59 & 3.7 & 0.86 & 28.7    \\

\midrule
\multicolumn{9}{c}{} \\
\midrule
\multirow{2}{*}{\textbf{Method}} &
\multicolumn{2}{c}{\textbf{arXiv}} &
\multicolumn{2}{c}{\textbf{DM Math}} &
\multicolumn{2}{c}{\textbf{HackerNews}} &
\multicolumn{2}{c}{\textbf{Average}} \\
\cmidrule(lr){2-3} \cmidrule(lr){4-5} \cmidrule(lr){6-7} \cmidrule(lr){8-9}
& ROC AUC & TPR@1\%FPR & ROC AUC & TPR@1\%FPR & ROC AUC & TPR@1\%FPR & ROC AUC & TPR@1\%FPR  \\
\midrule

AttenMIA (Non-Member Prefix)  & 0.82 & 33.4 & 0.96 & 89.6 & 0.62 & 6.6 & 0.77 & 35.1  \\
AttenMIA (Dropping Tokens)  & 0.82 & 31.6 & 0.96 & 90.6 & 0.61 & 10.5 & 0.77 & 35.9  \\
AttenMIA (Replacing Tokens)  & 0.81 & 36.2 & 0.96 & 89.5 & 0.60 & 9.0 & 0.77 & 35.8  \\

\bottomrule
\end{tabular}
}
\caption{ROC AUC and TPR@1\%FPR of AttenMIA under individual perturbation strategies across subsets of the MIMIR benchmark using the Pythia-1.4B model. Each yields similarly strong performance, indicating that each perturbation carries membership signals. Aggregating all perturbations provides the best overall results.}
\label{tab:ablation_results_perturbations}
\end{table*}

% ****************************************************************************************

\noindent \textbf{Analyzing the role of different layers in MIA.}
To study where memorization resides within the transformer, we analyze how different transformer layers contribute to membership inference. Using the perturbed attention features of the Pythia-1.4B model (24 layers) on the PubMed Central subset, we consider three settings: (i) features from a single layer, (ii) features grouped into lower, middle, and upper segments, and (iii) all layers combined.

\begin{figure}[h!]
    \centering
    \includegraphics[width=\linewidth, height=4cm]{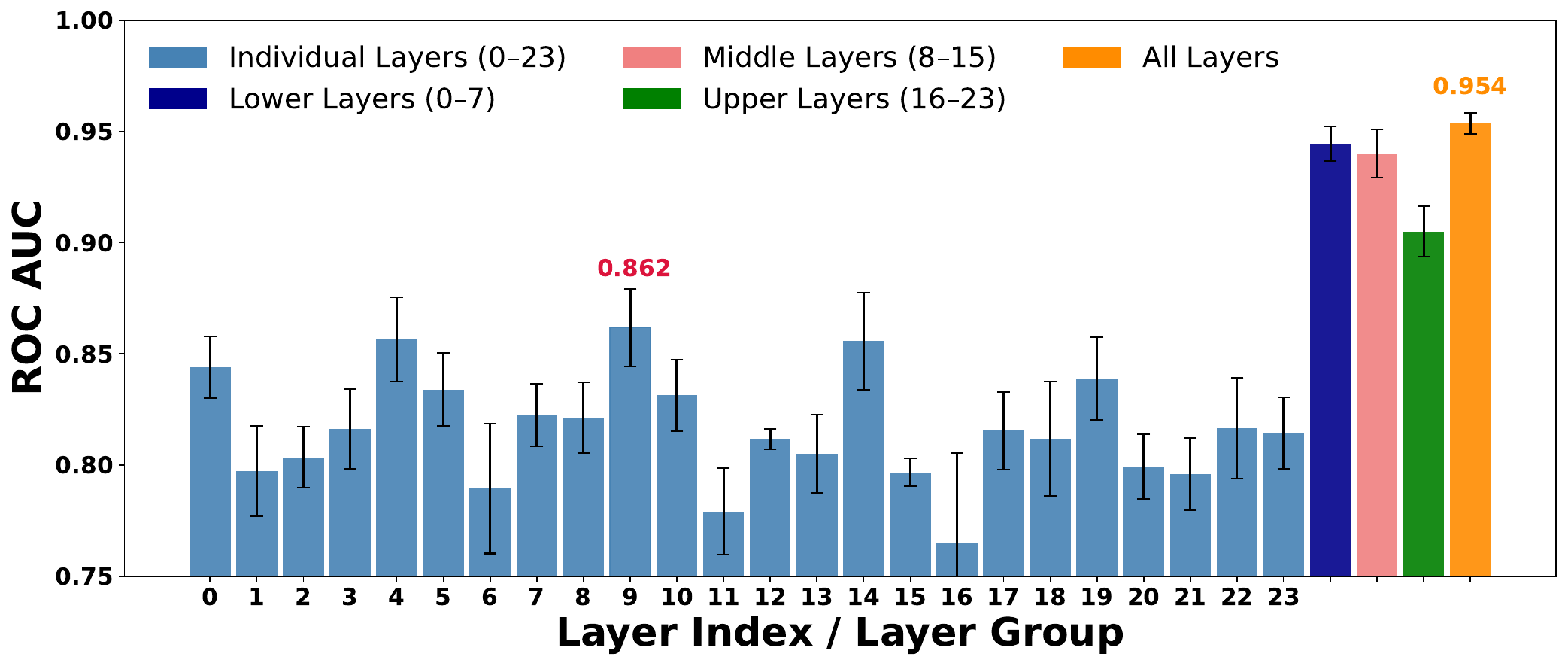}
    \caption{ROC AUC of AttenMIA when using perturbed attention features from different layers of Pythia-1.4B on the PubMed Central subset. We show results for individual layers (best layer \textcolor{D_red}{value} is highlighted). %, \textcolor{D_blue}{grouped lower/middle/upper layers}, and \textcolor{D_orange}{all layers combined}. 
    All layers contain exploitable membership signals, with performance improving when aggregating across groups and reaching its peak when using all layers together.}
    \label{fig:layer_analysis}
\end{figure}

Figure~\ref{fig:layer_analysis} shows that every layer contains meaningful membership information, though individual layers vary moderately in discriminative strength. Aggregating features within layer groups consistently improves performance over single layers, and using all 24 layers yields the highest ROC AUC. This suggests that memorization is not localized, and it is distributed across the layers and best captured when features from multiple depths are integrated.

\definecolor{D_red}{RGB}{214, 39, 40}

% ****************************************************************************************

\begin{figure}[h!]
    \centering
    \includegraphics[width=\linewidth]{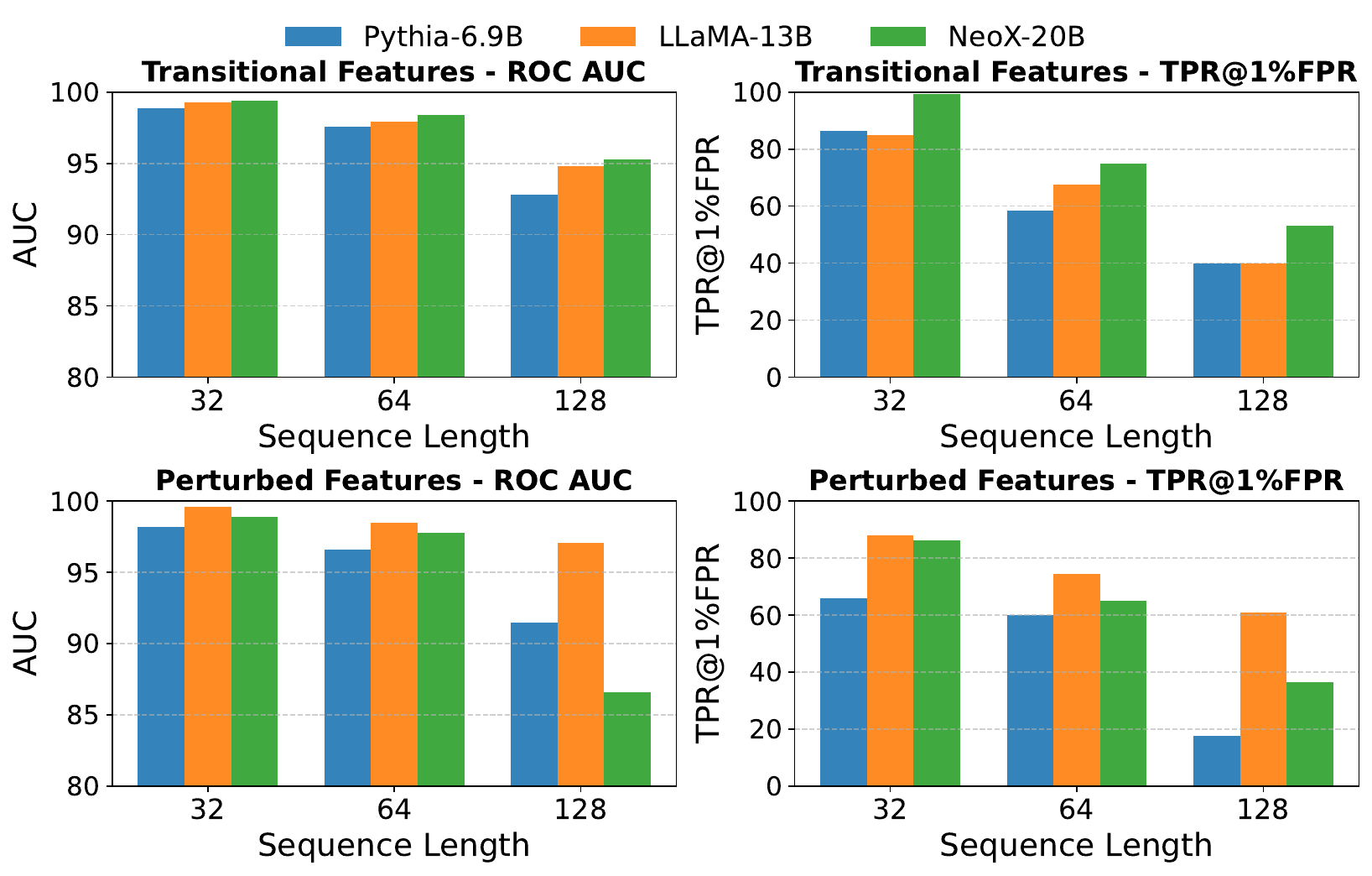}
    \caption{Effect of sequence length on AttenMIA performance. We report ROC AUC (left column) and TPR@1\%FPR (right column) for Pythia-6.9B, LLaMA-13B, and NeoX-20B using transitional (top) and perturbed (bottom) attention features. Shorter sequences exhibit stronger membership signals across models, while performance consistently degrades as sequence length increases.}
    \label{fig:length_effect}
\end{figure}

\noindent \textbf{Effect of sequence length on AttenMIA.}  
We examine how input sequence length influences membership inference. Using WikiMIA, we evaluate Pythia-6.9B, Llama2-13B, and NeoX-20B under two feature families, Transitional Features and Perturbed Features, across lengths 32, 64, and 128 tokens. Figure~\ref{fig:length_effect} represents two consistent trends.
First, shorter sequences produce stronger membership signals, especially for TPR@1\%FPR. With transitional features, NeoX-20B achieves nearly 100\% ROC AUC and TPR@1\%FPR at length 32; performance then gradually decreases as the sequence length increases. A similar pattern appears for perturbed features. This suggests that memorization signals are more concentrated and easier to extract in shorter contexts, whereas longer contexts introduce noise that obscures membership-specific patterns.
Second, despite this degradation, \textsc{AttenMIA} remains consistently effective across all models and sequence lengths, outperforming baseline attacks throughout (see Section~\ref{sec:main_results}). The larger performance drop in TPR@1\%FPR compared to AUC further indicates that the low-FPR setting is more sensitive to long contexts, an important consideration for real-world auditing.

% These findings underscore the importance of considering input length when auditing LLMs for memorization, as models may appear more robust when evaluated on longer sequences, even though shorter sequences expose stronger privacy risks.

% ------------------------------------------------------------------------------
% -----------------------------------------------------------------------------

\subsection{AttenMIA against Defenses}

\noindent \textbf{Training Data Deduplication.}
Prior work suggests that removing duplicated text from pre-training corpora can reduce memorization and thereby mitigate membership inference risks in LLMs~\cite{kandpal2022deduplicating, lee2021deduplicating}. The intuition is that duplicated data inflates memorization, so removing it should reduce vulnerability. This approach is generally more efficient than training-based defenses such as differential privacy, and it has been viewed as a practical mitigation strategy.

To evaluate whether deduplication weakens the signals exploited by \textsc{AttenMIA}, we test it on the Pythia-dedup family~\cite{biderman2023pythia}, which matches Pythia in architecture and size but is trained on a deduplicated version of the Pile~\cite{gao2020pile}. Pile is an open-source corpus used to train models like Pythia, composed of diverse sources such as Wikipedia, GitHub, arXiv, and PubMed. Table~\ref{tab:dedup_results_AUC} reports ROC AUC on the GitHub and HackerNews subsets across multiple model scales. Deduplication yields only negligible differences in attack performance, typically within 0.00–0.03 AUC. For instance, \textit{AttenMIA} achieves 0.99 ROC AUC on GitHub for all scales, regardless of deduplication. HackerNews shows a similarly minimal change, remaining around 0.70–0.71 ROC AUC.
% \nael{What is the effect of dedup on other attacks?  If it has substantial impact on them, but not attenMIA that is a good point.}

These results indicate that deduplication alone is insufficient to meaningfully reduce membership signals exploited by AttenMIA. Our results highlight the need for stronger or more principled defenses beyond deduplication to mitigate membership leakage in modern LLMs. Additional experiments are provided in Appendix~\ref{appendix: DEDUPLICATION}.

\begin{table*}[h!]
\centering
\resizebox{\textwidth}{!}{
\begin{tabular}{lcccccc}
\multicolumn{7}{c}{} \\
\toprule
\multirow{2}{*}{\textbf{Method}} &
\multicolumn{3}{c}{\textbf{GitHub (Pythia)}} &
\multicolumn{3}{c}{\textbf{HackerNews (Pythia)}} \\
\cmidrule(lr){2-4} \cmidrule(lr){5-7} 
& 1.4b & 2.8b & 6.9b & 1.4b & 2.8b & 6.9b  \\
\midrule

PPL attack  & 0.86 (0.00) & 0.87 (0.00) & 0.87 (-0.01) & 0.59 (0.00) & 0.60 (0.00) & 0.59 (-0.01)  \\

reference attack & 0.62 (-0.03) & 0.64 (-0.01) & 0.61 (-0.03) & 0.52 (0.00)  & 0.53 (0.00) & 0.52 (0.00)  \\

zlib attack & 0.85 (0.00) & 0.86 (0.00) & 0.86 (-0.01) & 0.59 (0.00) & 0.59 (0.00) & 0.59 (0.00) \\

neighborhood attack & 0.83 (0.00) & 0.82 (0.00) & 0.82 (-0.02) & 0.54 (0.00) & 0.54 (0.00) & 0.54 (-0.01)  \\

MIN-K\% PROB & 0.85 (-0.01) & 0.86 (0.00) & 0.87 (-0.01) & 0.56 (-0.01) & 0.58 (0.00) & 0.56 (-0.02)  \\

PETAL & 0.85 (0.00) & 0.86 (0.00) & 0.87 (-0.01) & 0.56 (-0.01) & 0.58 (0.00) & 0.56 (-0.02)  \\

\midrule
AttenMIA (Transitional)  & 0.90 (+0.02) & 0.92 (-0.01) & 0.88 (-0.01) & 0.61 (-0.01) & 0.59 (0.00) & 0.61 (+0.02)  \\
AttenMIA (Perturbed) & 0.99 (0.00)  & 0.99 (0.00) & 0.99 (-0.01) & 0.70 (0.00) & 0.71 (-0.01) & 0.71 (-0.01) \\

\bottomrule
\end{tabular}
}
\caption{ROC AUC of AttenMIA and baseline attacks on deduplicated Pythia models. Parentheses indicate the change relative to the corresponding non-deduplicated models. Deduplication produces only negligible differences across all architectures and datasets, showing that it does not meaningfully reduce membership inference vulnerability.}
\label{tab:dedup_results_AUC}
\end{table*}

%% file: sections/data_extraction.tex
\section{Training Data Extraction Using AttenMIA}
\label{sec: data_extraction}
% \pedram{Professor. This is the new version of data extraction using MIA, and compare it with the Carlini data extraction. Please read it and let me know if we need to discuss it.}
This section introduces a practical training-data extraction pipeline built on \textit{AttenMIA} and compares it to a strong baseline derived from prior work on identifying memorized generations in GPT-2~\cite{carlini2021extracting}. To ensure compatibility with established extraction protocols, we follow the standard setup used in Carlini et al~\cite{carlini2021extracting}. GPT-2 is prompted with short natural-language prefixes sampled from Internet text and evaluated on long continuations. Our goal is to determine, in an automated setting, which scoring strategies most effectively rank target model generations that correspond to training data. The baseline ranks candidates using likelihood-based and complexity-based metrics, whereas AttenMIA provides an alternative metric for ranking. For background on the extraction paradigm and its ethical considerations, read prior work~\cite{carlini2021extracting}.
% \nael{Previous sentence is unclear}

\subsection{Attack Setup}

We evaluate \textit{AttenMIA} in a controlled data extraction setting designed to mirror the protocol of Carlini et al.~\cite{carlini2021extracting}.  The goal is to determine whether AttenMIA can automatically identify memorized generations at scale, thereby removing the need for manual verification. 

\noindent \textbf{Attack setting.} 
Our target model is a publicly released GPT-2 model. For \textit{AttenMIA}, we assume that the adversary can access internal attention activations across all layers and heads to compute membership scores. Baseline extraction uses log-probabilities, perplexity, and sequence likelihood. As in prior extraction work, we don't assume access to the model's training corpus.

\noindent \textbf{Prefix corpus and generation procedure.}
Following established extraction protocols, we construct an evaluation corpus of short natural-language prefixes sampled from Internet-scale text (Common Crawl). Prefixes are 5–10 tokens long and are used to prompt GPT-2 to generate 256-token continuations under fixed decoding parameters. This process yields 14{,}000 prefix–continuation pairs. All scoring strategies, baseline, and AttenMIA operate on the same generated sequences to ensure fair comparison.

\noindent \textbf{Ground-truth references.}
For each sampled prefix, we retain the corresponding continuation from its source document as ground truth. These reference sequences allow automated, reproducible evaluation using ROUGE-L and other similarity metrics, replacing the manual verification and web-search stage used in~\cite{carlini2021extracting}. ROUGE-L evaluates the similarity between two sequences by computing the longest common subsequence (LCS) between them. Higher scores indicate stronger overlap and thus a greater likelihood of memorized or near-verbatim content. This design enables large-scale, controlled comparison of extraction scoring methods.
% \nael{Maybe need a description of ROUGE-L}

\noindent \textbf{Threat model.}
We consider an adversary with white-box access to an LLM but with no access to the training data. The attacker aims to rank generated continuations by their likelihood of having been included in the model's training set to identify which generations correspond to memorized samples.

\subsection{Scoring the Candidates}
We score each generated continuation using two families of scores: (i) baseline metrics introduced in Carlini et al.~\cite{carlini2021extracting}, and (ii) our proposed AttenMIA-based membership score. The goal is to assess which scoring functions most effectively rank generations that match their ground-truth references, so identify memorized content.

%\noindent \textbf{Baseline scoring metrics.} The baseline metrics follow the extraction methodology of~\cite{carlini2021extracting}, where each estimates the likelihood that a sample originates from the model's training distribution. They are defined as follows:

\noindent \textbf{Baseline scoring metrics.}
The baseline metrics follow the extraction methodology of~\cite{carlini2021extracting}, where each signal aims to estimate how ``training-like'' a generated continuation is. All baselines operate directly on the generated text, without using reference continuations. We implement the following scoring functions:

\begin{enumerate}

  \item \textbf{GPT-2 XL Perplexity (PPL-XL).} Measure the token-level surprise of the continuation under GPT-2 XL. Lower perplexity indicates the model finds the continuation more familiar and therefore potentially memorized.

  \item \textbf{S/XL Perplexity Ratio.} Memorized text is disproportionately easy for large models relative to small ones. Higher ratios indicate text that GPT-2 XL models but GPT-2 Small does not, a known memorization signature.

  \item \textbf{Lowercase/XL Ratio.} Measure sensitivity to casing by scoring the lowercased version of the continuation. Memorized content tends to retain distinctive casing patterns that disappear when lowercased.

  \item \textbf{Zlib Compression Score.} Compression length is used as a crude complexity proxy: simpler or more repetitive text compresses better. Lower values indicate text that is both low-surprisal and easily compressible.

\end{enumerate}

In the original extraction workflow, candidates are ranked by each metric and manually verified by human annotators to confirm whether they correspond to memorized text. In contrast, our evaluation replaces this costly step with automated reference-based similarity scoring, enabling systematic comparison at scale. 

\noindent \textbf{AttenMIA classifier score.}
For each generated continuation, we compute an \textit{AttenMIA} membership score derived from perturbation-based attention features. Specifically, we extract head-wise attention perturbed features (non-member prefix insertion, token deletions, and token replacements). These features are concatenated into a feature vector and fed into a multi-layer perceptron trained to classify members versus non-members. The resulting scalar output in [0,1] serves as a continuous ranking score, with larger values indicating stronger evidence of training-set membership.

\noindent \textbf{Evaluation metrics.}
To understand how well each scoring method identifies memorized generations, we measure how closely its ranking agrees with the actual similarity between a generated continuation and its ground-truth reference. We use ROUGE-L as an automated proxy for this similarity.

For a given scoring function \( s(\cdot) \), we first sort all generated samples by their score (higher scores indicating a stronger guess that the text is memorized). We then compute the \textit{Pearson correlation coefficient} \( r \) between the scores produced by \( s \) and the ROUGE-L similarity values.  Intuitively, this correlation captures how well a scoring method identifies the samples that truly match their original references. A higher correlation means that the metric is better aligned with actual memorization and, therefore, is more effective at ranking extracted candidates.

\subsection{Experimental Results and Analysis}

To evaluate the effectiveness of our proposed approach, we measure how well each metric aligns with true memorization quality. For each generated continuation, we compute the \textit{Pearson correlation coefficient} ($r$) between each metric's score and the ROUGE-L similarity to its ground-truth reference. A higher correlation indicates that the scoring function more accurately ranks generations according to their likelihood of reflecting training data.

\noindent \textbf{Evaluation protocol.} 
We analyze 2{,}000 samples: the top-$1{,}000$ and bottom-$1{,}000$ generations ranked by ROUGE-L. This selection covers both highly faithful and weakly related continuations, enabling a balanced assessment of how well each scoring method captures linear trends in memorization strength. For each scoring function~$s(\cdot)$, we compute Pearson's~$r$ over this dataset.

\noindent \textbf{Results.} 
Table~\ref{tab:data_extraction_pearson} reports the correlations. The \textit{AttenMIA} classifier achieves the strongest alignment with ground-truth similarity ($r{=}0.48$), substantially outperforming all baseline metrics. The best performing baseline, the Zlib/XL Ratio, reaches only $r{=}0.32$. These results show that attention-based perturbation features encode a more direct signal of training-data familiarity than surface-level measures such as perplexity or compression ratios.

\noindent \textbf{Examples.}
Figure~\ref{fig: DE_examples} illustrates two examples where \textit{AttenMIA} assigns high membership scores to continuations that exhibit strong ROUGE-L overlap with their ground-truth references. In both cases, the model reproduces long spans of the original text. These examples illustrate how \textit{AttenMIA} serves as a reliable indicator of memorized continuations without requiring human inspection or manual filtering.

\noindent \textbf{Interpretation.}
A higher correlation with ROUGE-L implies that a scoring function assigns a greater weight to generations that resemble training data. The superior performance of \textit{AttenMIA} suggests that attention-based features carry signals that are not reflected in output-level signals.

\noindent \textbf{Ethical considerations.}
We use only publicly available models and non-sensitive Internet text in our experiments. Our goal with \textit{AttenMIA} is to shed light on how and when large language models memorize data—not to enable the extraction of private or personal information. By better understanding these behaviors, we aim to help developers build models that are safer and more respectful of user privacy, and to provide tools that support responsible auditing of memorization in deployed systems.

\begin{table}[h!]
\centering
\small
\resizebox{\columnwidth}{!}{%
\begin{tabular}{l|c|cc}
\toprule
\textbf{Metric} & \textbf{Pearson $r$} & \textbf{Interpretation}  \\
\midrule
ratio\_zlib\_xl & 0.321 & Moderate correlation  \\
ratio\_lower\_xl & 0.070 & Weak correlation \\
ratio\_s\_xl & 0.048 & Very weak correlation \\
zlib\_entropy & $-0.092$ & Near zero correlation \\
ppl\_xl & $-0.206$ & Negative correlation \\
ppl\_lower & $-0.238$ & Negative correlation \\
ppl\_s & $-0.256$ & Negative correlation \\
\midrule
\textbf{AttenMIA} & \textbf{0.480} & Strong correlation \\
\bottomrule
\end{tabular}}
\caption{Pearson correlation ($r$) between each scoring metric and ROUGE-L similarity. A higher magnitude of $r$ indicates stronger alignment with memorization quality. \textit{AttenMIA} achieves the highest correlation, substantially outperforming all baseline metrics.
}
\label{tab:data_extraction_pearson}
\end{table}

% \begin{figure}[h!]
%     \centering
%     \includegraphics[width=\columnwidth]{img/Extracted.png}
%     \caption{Examples of top-ranked generations identified by \textit{AttenMIA}. Both samples exhibit high ROUGE-L similarity and strong membership confidence. The first example reproduces a near-verbatim country list from a public encyclopedic source, while the second recreates website content from Wikidot. \pedram{I will replace this figure with a better version as soon as possible.}
%     }
%     \label{fig: DE_examples}
% \end{figure}

\begin{figure}[h!]
\centering

\begin{tcolorbox}[
    colback=gray!3,
    colframe=black!20,
    arc=2pt,
    left=1mm,
    right=1mm,
    boxsep=1.5mm,
    % borderline between segments={0.3mm}{0pt}{black!20,dashed}
]

\textbf{Prefix:} Egypt El Salvador Equatorial Guinea Eritrea

\medskip
\textbf{Extraction:} Estonia Ethiopia Falkland Islands (Malvinas) Fare Islands
Fiji Finland France French Guiana French Polynesia French Southern Territories
Georgia Germany Gibraltar~\ldots

\medskip
\textbf{ROUGE-L:} {\color{blue!60}0.901} \quad
\textbf{AttenMIA:} {\color{purple!70}1.0}

\tcblower

\textbf{Prefix:} how this page has evolved in the past.

\medskip
\textbf{Extraction:} ``If you want to discuss contents of this page this is the easiest
way to do it. View and manage~\ldots\ General Wikidot.com documentation and
help section. Wikidot.com Terms of Service~\ldots''

\medskip
\textbf{ROUGE-L:} {\color{blue!60}0.843} \quad
\textbf{AttenMIA:} {\color{purple!70}0.783}

\end{tcolorbox}

\caption{Examples of top-ranked generations identified by \textit{AttenMIA}.  
Both generations show high ROUGE-L similarity and strong membership confidence.}
\label{fig:DE_examples}
\label{fig: DE_examples}
\end{figure}

%% file: sections/related_work.tex
\section{Related Work}

In this section, we present the works most related to our contributions in this paper.  We organize these in three groups: (1) Membership inference attacks; (2) training data extraction attacks, and (3) work leveraging attention for attacks or safety analysis in LLMs. \\

\noindent \textbf{Membership Inference Attacks (MIAs):} Membership inference attacks aim to determine whether a specific example was present in a model's training set~\cite{shokri2017membership}. Early methods relied primarily on loss or likelihood-based information. Yeom et al.~\cite{yeom2018privacy} utilize the model loss-based signal, while Carlini et al.\cite{carlini2021extracting} observe low-perplexity generations as potential indicators of memorization. RECALL~\cite{xie2024recall} observes the change in conditional log-likelihood to determine the member.

Reference-based MIAs use shadow models to approximate target behavior. LiRA~\cite{carlini2022membership} and Truth Serum~\cite{tramer2022truth} achieve strong performance but require training dozens of shadow models, making them computationally costly for LLMs.

More recent approaches avoid shadow training: neighborhood perturbation~\cite{mattern2023membership}, and semantic MIAs~\cite{mozaffari2024semantic} operate directly on the target model without auxiliary training. PETAL~\cite{he2025towards} leverages token-level semantic similarity to approximate output probabilities and subsequently calculate the perplexity. Chen et al.~\cite{chen2024method} and Chaudhari et al.~\cite{chaudhari2023chameleon} poison the dataset and then observe the model output only to determine the member samples. Moreover, with access to model internal states~\cite{song2019privacy}, attackers can exploit model gradients~\cite{nasr2019comprehensive} or representations~\cite {song2019privacy} to leak member samples, as well as amplify the membership leakage of fine-tuned LLMs~\cite{mireshghallah2022empirical, fu2024membership}. 

Despite this progress, no prior work systematically examines attention patterns as a signal for membership inference in LLMs. To our knowledge, \textit{AttenMIA} is the first framework to extract attention-based features, enabling high MIA accuracy. \\

\noindent \textbf{Data Extraction Attacks:} A successful training data extraction attack checks the model's output against the original training data for an exact~\cite{yu2023bag,carlini2023quantifying,lee2023language,carlini2019secret,carlini2021extracting,panaitescu2025poisonedparrot} or near-exact (e.g., 50 tokens) match~\cite{nasr2023scalable}. Carlini et al.~\cite{carlini2019secret} first quantified the data extraction attack by systematically prompting a model in which an adversary recovers unique secrets (e.g., credit card numbers) that were present in the training set. In the subsequent work, Carlini et al.~\cite{carlini2021extracting} prompted the model with randomly sampled prefixes, generating millions of completions and then checking each against the original training corpus, thus extracting hundreds of verbatim text sequences from GPT-2 training data. Nasr et al.~\cite{nasr2023scalable} extracted several megabytes of ChatGPT's training data at a cost of only a few hundred USD in API queries. Mamun et al.~\cite{al2023deepmem} and Yu et al.~\cite{yu2023bag} showed that with careful prompt engineering and decoding parameters, one can significantly boost the accuracy of training data extraction attacks compared to the baseline methods. Another line of research surveys on extracting training data from a given pre-trained language model~\cite{ishihara2023training}. For example, Carlini et al.~\cite{carlini2023quantifying} extracted a subset of training data while prompting the pretrained GPT-Neo model family. Lee et al.~\cite{lee2023language} found that the top-k and top-p sampling strategy leads to extracting more training data. There are also some works which show that duplication of training data significantly influences memorization and generates the training data verbatim in its responses~\cite{kandpal2022deduplicating,carlini2023quantifying}. \\

\noindent \textbf{Exploiting attention for attacks and safety:} In general, adversarial attacks on transformer-based systems have largely not exploited attention, despite the emerging understanding of its critical role in managing information flow in these models~\cite{vaswani2017attention,ben2024attend,zheng2024attentionheadslargelanguage}.  We are not aware of any privacy-based attacks exploiting attention, but some works have explored using attention for jailbreaking LLMs~\cite{zaree2025attention,pandya2025iattentionbreakingfinetuning} or alternatively for detecting such attacks~\cite{pu2025feintattackattentionbasedstrategies}.

%% file: sections/conclusion.tex
\section{Conclusion}
In this work, we introduced \textsc{AttenMIA}, a new framework that uses attention-based features as a signal for membership inference in large language models. By designing both transitional and perturbation-based attention features, we showed that internal attention
dynamics, often viewed as a pathway toward interpretability, also carry strong indicators of memorization. 

Across multiple open-weight LLMs, datasets, and evaluation settings, \textsc{AttenMIA} consistently outperformed existing membership inference attacks. Notably, it excelled in the low-FPR regime that is critical for practical privacy auditing. Our layer- and head-level analysis further revealed where and how memorization emerges inside the transformers, providing new insight into the structure of training-data leakage.

Taken together, our findings highlight an important tension in modern LLM design: mechanisms introduced to make models more interpretable can also introduce new privacy risks. We hope these results motivate future work on privacy-preserving training strategies, as well as the development of auditing tools that leverage attention pathways responsibly. Ultimately, \textsc{AttenMIA} demonstrates that attention is not only a lens for understanding model behavior, but also a powerful signal for measuring and mitigating unintended memorization in large-scale language models.

%% file: sections/acknowledgments.tex
% use section* for acknowledgment
\iffalse 
\ifCLASSOPTIONcompsoc
  % The Computer Society usually uses the plural form
  \section*{Acknowledgments}
\else
  % regular IEEE prefers the singular form
  \section*{Acknowledgment}
\fi

We thank the anonymous reviewers for their constructive feedback, which helped improve the clarity and quality of this work. We are also grateful to the maintainers of the open-source models and datasets used in our experiments, without which this research would not have been possible.

% This work was supported in part by [Funding Agency/Grant Number], and computational resources were provided by GCP. 
\fi

%% file: sections/appendix.tex
\appendix

\section{Background on Transformer Attention}
\label{app:attention}

\subsection{Additional Results on MIMIR benchmark}
\label{sec: big_table}

Table~\ref{tab:main_results_mimir_TPR} shows the TPR@1\%FPR of baselines and AttenMIA on the MIMIR benchmark.

% ------------------------------------------------------------------------------
% ------------------------------------------------------------------------------

\begin{table*}[h!]
\centering
\resizebox{\textwidth}{!}{
\begin{tabular}{lcccccccccccc}
\multicolumn{13}{c}{} \\
\toprule
\multirow{2}{*}{\textbf{Method}} &
\multicolumn{3}{c}{\textbf{Wikipedia (Pythia)}} &
\multicolumn{3}{c}{\textbf{GitHub (Pythia)}} &
\multicolumn{3}{c}{\textbf{Pile CC (Pythia)}} &
\multicolumn{3}{c}{\textbf{PubMed (Pythia)}} \\
\cmidrule(lr){2-4} \cmidrule(lr){5-7} \cmidrule(lr){8-10} \cmidrule(lr){11-13}
& 1.4b & 2.8b & 6.9b & 1.4b & 2.8b & 6.9b & 1.4b & 2.8b & 6.9b & 1.4b & 2.8b & 6.9b \\
\midrule

PPL attack & 4.8 & 4.0 & 3.2 & 35.6 & 43.2 & 48.8 & 4.0 & 4.4 & 5.6 & 4.8 & 4.8 & 2.0  \\

reference attack & 3.6 & 0.8 & 1.6 & 15.2 & 12.8 & 9.6 & 3.2 & 2.8 & 3.2 & 2.0 & 1.2 & 1.6 \\

zlib attack  & 2.0 & 2.4 & 2.4 & 27.6 & 31.2 & 36.8 & \textbf{4.4} & 5.6 & \textbf{6.4} & 5.2 & 3.6 & 3.2 \\

neighborhood attack  & 0.4 & 0.8 & 1.2 & 32.0 & 26.0 & 27.2 & 3.2 & 3.2 & 5.2 & 8.4 & 7.2 & 5.2 \\

MIN-K\% PROB  & 2.0 & 0.4 & 2.0 & 23.6 & 31.6 & 28.4 & 3.6 & 2.8 & 4.4 & 1.6 & 1.2 & 1.6 \\

WS attack  & 2.8 & 1.6 & 4.4 & 13.6 & 17.2 & 21.6 & 1.2 & 0.4 & 2.4 & 0.4 & 1.2 & 0.8 \\

RS attack  & 3.2 & 2.0 & 3.6 & 28.8 & 32.0 & 35.6 & 0.0 & 2.4 & 0.4 & 0.8 & 2.0 & 2.8   \\

BT attack & 6.0 & 2.8 & 3.6 & 22.8 & 25.2 & 25.2 & 2.4 & 2.4 & 2.0 & 3.6 & 4.4 & 4.0 \\

PETAL  & 3.2 & 2.8 & 4.0 & 31.2 & 33.6 & 44.4 & 4.0 & 4.4 & 5.6 & 6.0 & 4.0 & 4.4 \\

RECALL & 10.7 & 11.7 & 13.6 & 48.8 & 49.6 & 52.7 & 2.9 & 4.0 & 4.9 & 13.8 & 19.4 & 24.4 \\

\midrule

AttenMIA (Transitional)  & 16.7 & 7.6 & 7.7 & 60.7 & 73.8 & 60.1 & 4.2 & 5.0 & 5.0 & 31.6 & 25.9 & 24.8  \\

AttenMIA (Perturbed)  & \textbf{22.7} & \textbf{20.0} & \textbf{20.7} & \textbf{91.3} & \textbf{95.6} & \textbf{95.4} & 3.7 & \textbf{7.4} & 6.3 & \textbf{42.8} & \textbf{67.3} &  \textbf{63.7} \\

\midrule
\multicolumn{13}{c}{} \\
\midrule
\multirow{2}{*}{\textbf{Method}} &
\multicolumn{3}{c}{\textbf{arXiv (Pythia)}} &
\multicolumn{3}{c}{\textbf{DM Math (Pythia)}} &
\multicolumn{3}{c}{\textbf{HackerNews (Pythia)}} &
\multicolumn{3}{c}{\textbf{Average (Pythia)}} \\
\cmidrule(lr){2-4} \cmidrule(lr){5-7} \cmidrule(lr){8-10} \cmidrule(lr){11-13}
& 1.4b & 2.8b & 6.9b & 1.4b & 2.8b & 6.9b & 1.4b & 2.8b & 6.9b & 1.4b & 2.8b & 6.9b \\
\midrule

PPL attack   & 10.4 & 6.8 & 10.4 & 51.7 & 34.8 & 42.7 & 4.8 & 6.0 & 5.6 & 16.6 & 14.9 & 16.9  \\

reference attack & 2.4 & 2.0 & 1.6 & 49.4 & 50.6 & 47.2 & 1.2 & 1.2 & 1.2 & 11.0 & 10.2 & 
 9.4  \\

zlib attack  & 10.8 & 10.8 & 12.8 & 41.6 & 23.6 & 38.2 & 4.8 & 4.4 & 5.2 & 13.8 & 11.7 & 15.0 \\

neighborhood attack & 14.8 & 12.4 & 12.0 & 2.2 & 1.1 & 4.5 & 0.8 & 2.4 & 1.2 & 8.8 & 7.6 & 8.1 \\

MIN-K\% PROB  & 4.4 & 1.6 & 4.4 & 21.3 & 19.1 & 12.4 & 3.2 & 4.8 & 4.8 & 8.5 & 8.8 & 8.3 \\

WS attack & 1.6 & 1.2 & 0.0 & 0.0 & 2.2 & 1.1 & 0.4 & 1.2 & 2.0 & 2.9 & 3.6 & 4.6 \\

RS attack  & 1.2 & 0.8 & 0.8 & 0.0 & 2.2 & 0.0 & 0.4 & 0.8 & 0.4 & 4.9 & 6.0 & 6.2 \\

BT attack & 0.4 & 2.8 & 1.6 & 1.1 & 3.4 & 3.4 & 0.0 & 0.4 & 0.0 & 5.2 & 5.9 & 5.7 \\

PETAL & 5.2 & 4.8 & 6.4 & 33.7 & 37.1 & 29.2 & 3.6 & 4.4 & 6.8 & 12.4 & 13.0 & 14.4  \\

RECALL  & 15.6 & 15.6 & 17.4 & 44.2 & 31.2 & 22.1 & 1.7 & 2.7 & 4.9 & 19.7 & 19.2 & 20.0 \\

\midrule

AttenMIA (Transitional)  & 31.6 & 28.8 & 35.2 & 86.7 & 87.8 & 85.6 & 7.4 & 4.8 & 5.1 & 34.1 & 33.4 & 31.9  \\

AttenMIA (Perturbed)  & \textbf{42.3} & \textbf{50.9} & \textbf{55.4} & \textbf{98.9} & \textbf{98.8} & \textbf{99.3} & \textbf{13.8} & \textbf{13.0} & \textbf{18.9} & \textbf{41.6} & \textbf{50.4} & \textbf{51.4}   \\

\bottomrule
\end{tabular}
}
\caption{TPR@1\%FPR of membership inference attacks across seven datasets and three Pythia model sizes (1.4B, 2.8B, 6.9B). AttenMIA (Transitional and Perturbed) achieves the highest TPR across nearly all dataset–model configurations, substantially outperforming prior attacks in the low-FPR regime. Bold values denote the best result per configuration.}
\label{tab:main_results_mimir_TPR}
\end{table*}

\subsection{Effects of Independent and Cumulative Masking}

\begin{figure*}[h!]
    \centering

    % --- Subfigure (a) ---
    \begin{subfigure}{0.48\linewidth}
        \centering
        \includegraphics[width=\linewidth]{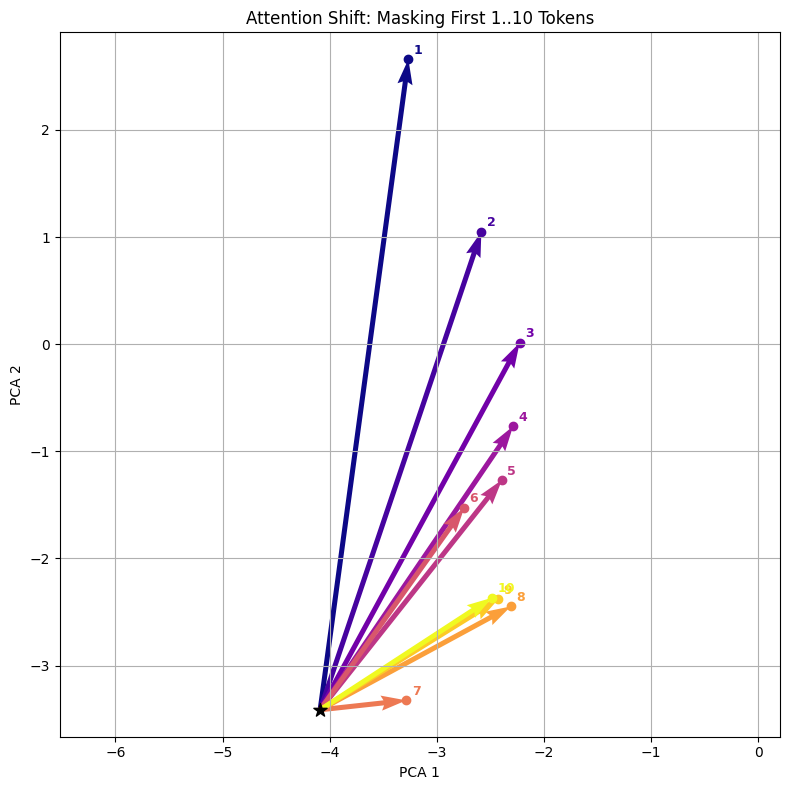}
        \caption{Independent masking: each token is masked individually.}
        \label{fig:independent_masking}
    \end{subfigure}
    \hfill
    % --- Subfigure (b) ---
    \begin{subfigure}{0.48\linewidth}
        \centering
        \includegraphics[width=\linewidth]{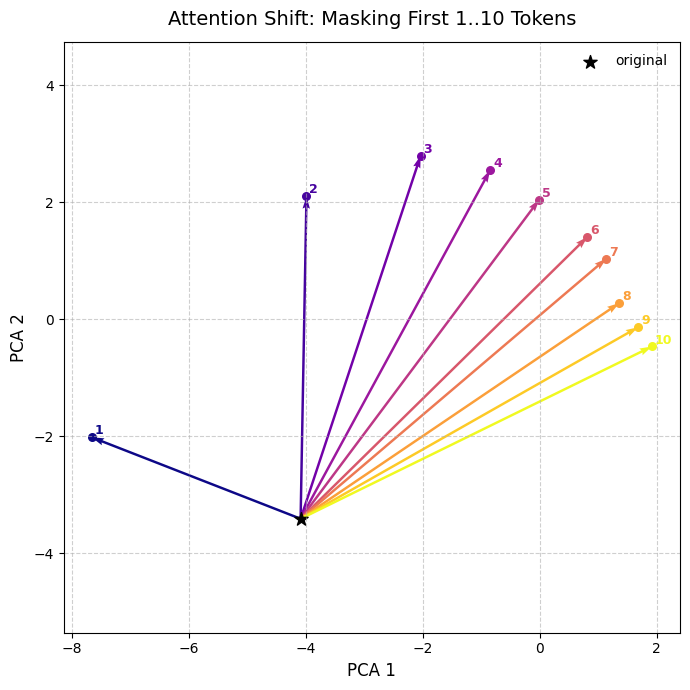}
        \caption{Cumulative masking: first $k$ tokens progressively masked.}
        \label{fig:cumulative_masking}
    \end{subfigure}

    \caption{Comparison of attention shifts under two masking strategies.}
    \label{fig:masking_comparison}
\end{figure*}

We conducted two complementary masking experiments to examine how attention patterns shift when input tokens are obscured. The first experiment masks tokens individually, while the second applies cumulative masking of the first $k$ tokens. Both sets of results were projected into two-dimensional PCA \cite{van2009dimensionality} space for interpretability, and the corresponding plots are shown in Figures~\ref{fig:independent_masking} and~\ref{fig:cumulative_masking}. \\

\noindent \textbf{Independent masking.} In the independent masking condition (Figure~\ref{fig:independent_masking}), each token is masked in isolation, allowing us to probe its specific contribution to the attention distribution. The arrows in the PCA space represent the shift relative to the original unmasked sequence. We observe that masking the earliest tokens results in large displacements, with arrows extending farther away from the origin. This indicates that the model relies heavily on the very first tokens to anchor its global representation. In contrast, masking later tokens produces smaller shifts that point in diverse directions. These shorter arrows suggest that deeper tokens influence local refinements rather than driving the overall attention structure. Thus, the early prefix has a disproportionately strong effect, while subsequent tokens contribute incrementally and in more heterogeneous ways. \\

\noindent \textbf{Cumulative masking.} In the cumulative masking condition (Figure~\ref{fig:cumulative_masking}), the first $k$ tokens are progressively masked together ($1{:}i$ for $i=1, \dots, k$). Interestingly, the pattern here contrasts with the independent case. Masking only the first token produces a relatively small shift, but as more tokens are aggregated, the displacement grows larger and moves in a consistent direction. This trend continues until approximately six or seven tokens are masked, at which point the vectors reach their maximum length, signifying the strongest deviation from the original representation. Beyond this point, the arrows gradually shorten again, but converge to a different region in PCA space. This suggests that the combined influence of the first several tokens is greater than any single token alone, and that the model’s representation undergoes its largest structural change when the prefix up to token 6-7 is removed. After this threshold, additional masking alters the representation less drastically, possibly because the prefix information has already been effectively erased. \\

\noindent \textbf{Summary.} Together, the two experiments highlight complementary aspects of transformer attention. Independent masking emphasizes the dominance of the first few tokens in shaping the global representation, while cumulative masking reveals a compounding effect in which the joint removal of multiple prefix tokens produces the greatest shift. These observations underscore the dual role of initial tokens: both as powerful individual anchors and as a collective foundation that, when removed, forces the model to reorganize its attention distribution.

% \begin{figure*}[h!]
%     \centering
%     % --- Left figure ---
%     \begin{subfigure}[t]{0.48\textwidth}
%         \centering
%         \includegraphics[width=\linewidth]{img/independent_masking.png}
%         \caption{Independent masking: each token is masked individually.}
%         \label{fig:independent_masking}
%     \end{subfigure}%
%     \hfill
%     % --- Right figure ---
%     \begin{subfigure}[t]{0.48\textwidth}
%         \centering
%         \includegraphics[width=\linewidth]{img/cumulative_masking.png}
%         \caption{Cumulative masking: first $k$ tokens progressively masked.}
%         \label{fig:cumulative_masking}
%     \end{subfigure}
%     %
%     \caption{Comparison of attention shifts under two masking strategies.}
%     \label{fig:masking_comparison}
% \end{figure*}

% \begin{figure}[h!]
%     \centering
%     % --- Subfigure (a) ---
%     \begin{subfigure}{\linewidth}
%         \centering
%         \includegraphics[width=\linewidth]{img/independent_masking.png}
%         \caption{Independent masking: each token is masked individually.}
%         \label{fig:independent_masking}
%     \end{subfigure}

%     \vspace{1em} % Optional space between the two subfigures

%     % --- Subfigure (b) ---
%     \begin{subfigure}{\linewidth}
%         \centering
%         \includegraphics[width=\linewidth]{img/cumulative_masking.png}
%         \caption{Cumulative masking: first $k$ tokens progressively masked.}
%         \label{fig:cumulative_masking}
%     \end{subfigure}

%     \caption{Comparison of attention shifts under two masking strategies.}
%     \label{fig:masking_comparison}
% \end{figure}

\subsection{Training Data Deduplication: Additional Results}
\label{appendix: DEDUPLICATION}

A common defense strategy against membership inference attacks is \emph{training data deduplication}, where duplicate samples in the pre-training corpus are removed to reduce spurious memorization. Prior studies have suggested that deduplication could mitigate privacy risks in large language models while being more efficient than training-based methods such as differential privacy. To test this hypothesis, we evaluate AttenMIA and baseline attacks on the deduplicated Pythia family (Pythia-dedup), which maintains the same parameter sizes as the original models but is trained on a deduplicated version of the Pile dataset.  

Table~\ref{tab:dedup_results_TPR} presents the TPR@1\%FPR results on GitHub and HackerNews subsets. Across both benchmarks, we observe that deduplication only marginally reduces the attack performance. For example, AttenMIA with perturbation features still achieves very strong detection rates (up to 97.3\% on GitHub and 16.5\% on HackerNews), with changes relative to the non-deduplicated models being negligible in most cases. These findings indicate that coarse-grained deduplication is insufficient to remove memorization signals exploitable by attention-based attacks. This aligns with prior work suggesting that fine-grained deduplication may be necessary but comes at the cost of utility, since valuable knowledge may also be discarded. Overall, our results highlight that more robust defenses are needed beyond deduplication to effectively mitigate membership inference risks.

\begin{table*}[h!]
\centering
\resizebox{\textwidth}{!}{
\begin{tabular}{lcccccc}
% \toprule
\multicolumn{7}{c}{} \\
\toprule
\multirow{2}{*}{\textbf{Method}} &
\multicolumn{3}{c}{\textbf{GitHub}} &
\multicolumn{3}{c}{\textbf{HackerNews}} \\
\cmidrule(lr){2-4} \cmidrule(lr){5-7} 
& Pythia-1.4b & Pythia-2.8b & Pythia-6.9b & Pythia-1.4b & Pythia-2.8b & Pythia-6.9b  \\
\midrule

PPL attack  & 36.8\% (+1.2\%) & 44.4\% (+1.2\%) & 52.0\% (+3.2\%) & 4.8\% (+0.0\%) & 6.0\% (+0.0\%) & 5.2\% (-0.4\%) \\

reference attack & 12.4\% (-2.8\%) & 11.2\% (-1.6\%) & 10.0\% (+0.4\%) & 0.4\% (-0.8\%) & 1.2\% (+0.0\%) & 1.6\% (+0.4\%) \\

zlib attack & 29.6\% (+2.0\%) & 30.0\% (-1.2\%) & 40.0\% (+3.2\%) & 4.4\% (-0.4\%) & 4.4\% (+0.0\%) & 5.2\% (+0.0\%) \\

neighborhood attack & 24.0\% (-8.0\%) & 28.4\% (+2.4\%) & 28.4\% (+1.2\%) & 6.0\% (+5.2\%) & 3.6\% (+1.2\%) & 3.2\% (+2.0\%) \\

MIN-K\% PROB & 25.2\% (+1.6\%) & 30.8\% (-0.8\%) & 29.2\% (+0.8\%) & 2.4\% (-0.8\%) & 3.6\% (-1.2\%) & 4.4\% (-0.4\%) \\

PETAL & 40.4\% (+9.2\%) & 38.0\% (+4.4\%) & 40.8\% (-3.6\%) & 4.4\% (+0.8\%) & 5.2\% (+0.8\%) & 6.0\% (-0.8\%) \\

\midrule

AttenMIA (Transitional)  & 68.7\% (+8.0\%) & 66.5\% (-7.3\%) & 43.0\% (-17.1\%) & 8.2\% (+0.8\%) & 5.0\% (+0.2\%) & 3.7\% (-1.4\%)  \\
AttenMIA (Perturbed) & 92.8\% (+1.5\%) & 97.3\% (+1.7\%) & 94.7\% (-0.7\%) & 15.1\% (+1.3\%) & 16.5\% (+3.5\%) & 16.5\% (-2.4\%)  \\

\bottomrule
\end{tabular}
}
\caption{TPR@1\%FPR results of AttenMIA and baselines on deduplicated Pythia models. Numbers in parentheses denote the change relative to the non-deduplicated Pythia counterparts.}
\label{tab:dedup_results_TPR}
\end{table*}

\subsection{ROC and PR Curves of Main Experiments}
\label{appendix:roc_pr_main}

To complement the main experimental results reported in Section~\ref{sec:main_results}, we provide the full ROC and Precision-Recall (PR) curves for \textit{AttenMIA} evaluated on the MIMIR benchmark. While the main text primarily reports aggregate metrics such as ROC AUC and TPR@1\%FPR, visualizing the curves offers additional insights into the stability and reliability of our attack across different thresholds.

\begin{figure*}[h!]
    \centering

    \subfloat[arXiv]{\includegraphics[width=0.45\linewidth]{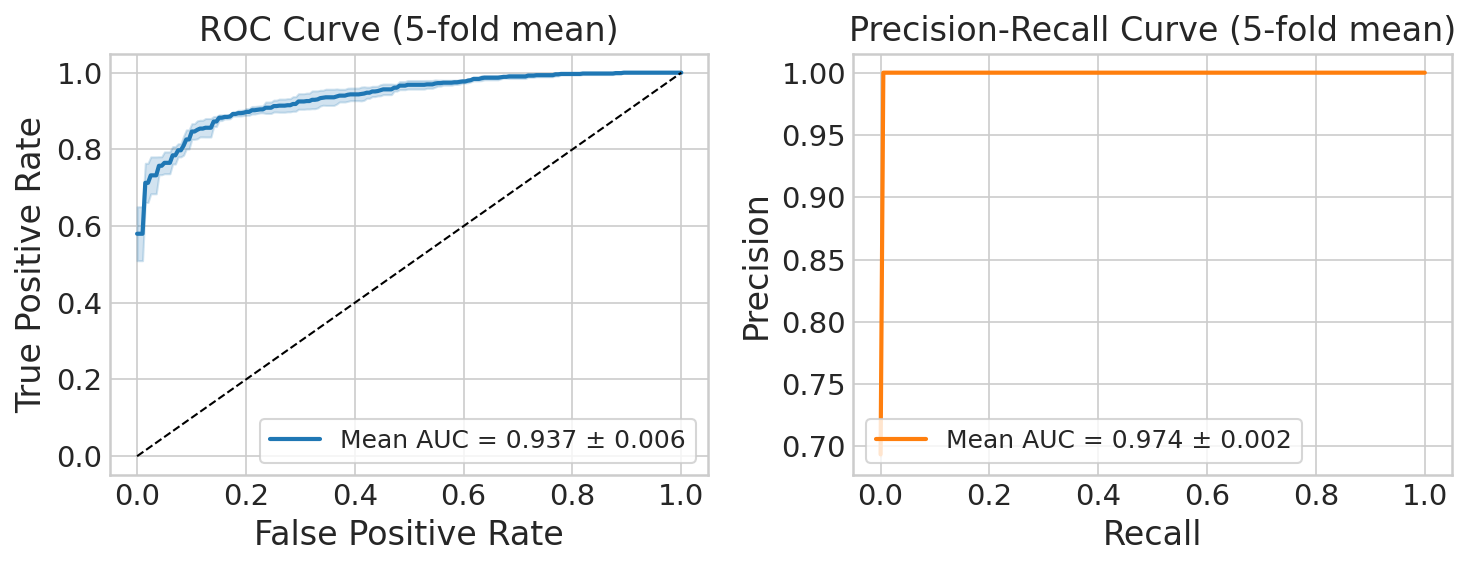}}
    \subfloat[DM Math]{\includegraphics[width=0.45\linewidth]{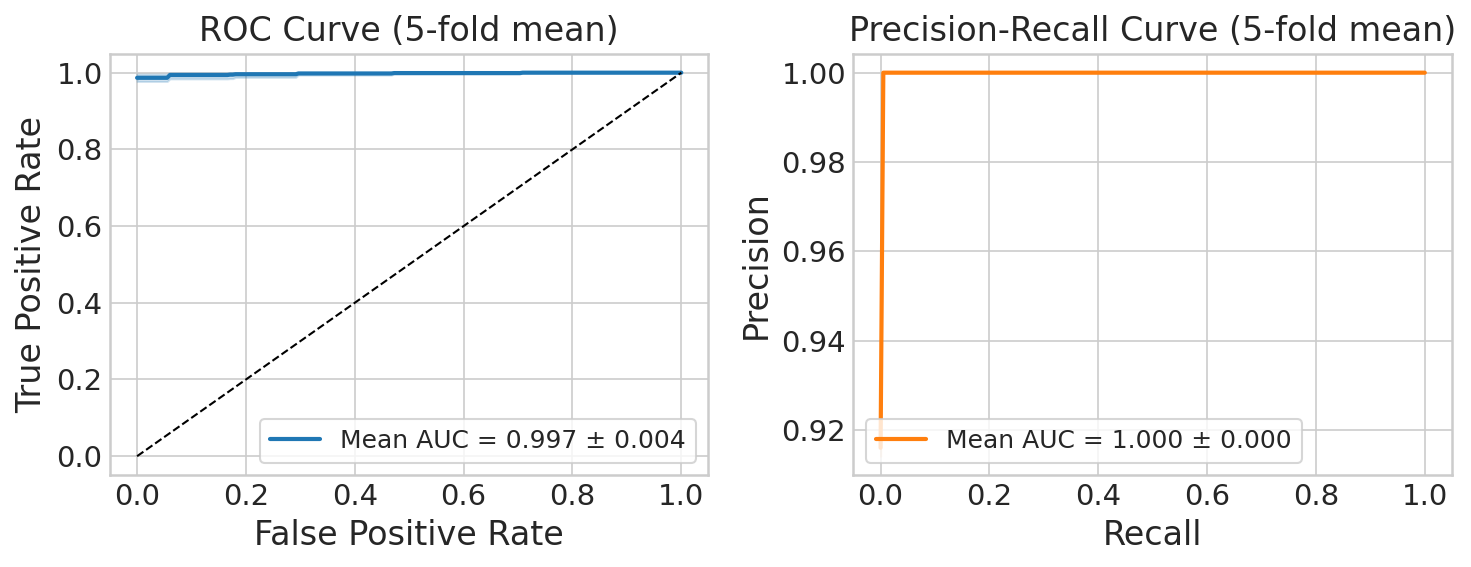}} \\
    
    \subfloat[GitHub]{\includegraphics[width=0.45\linewidth]{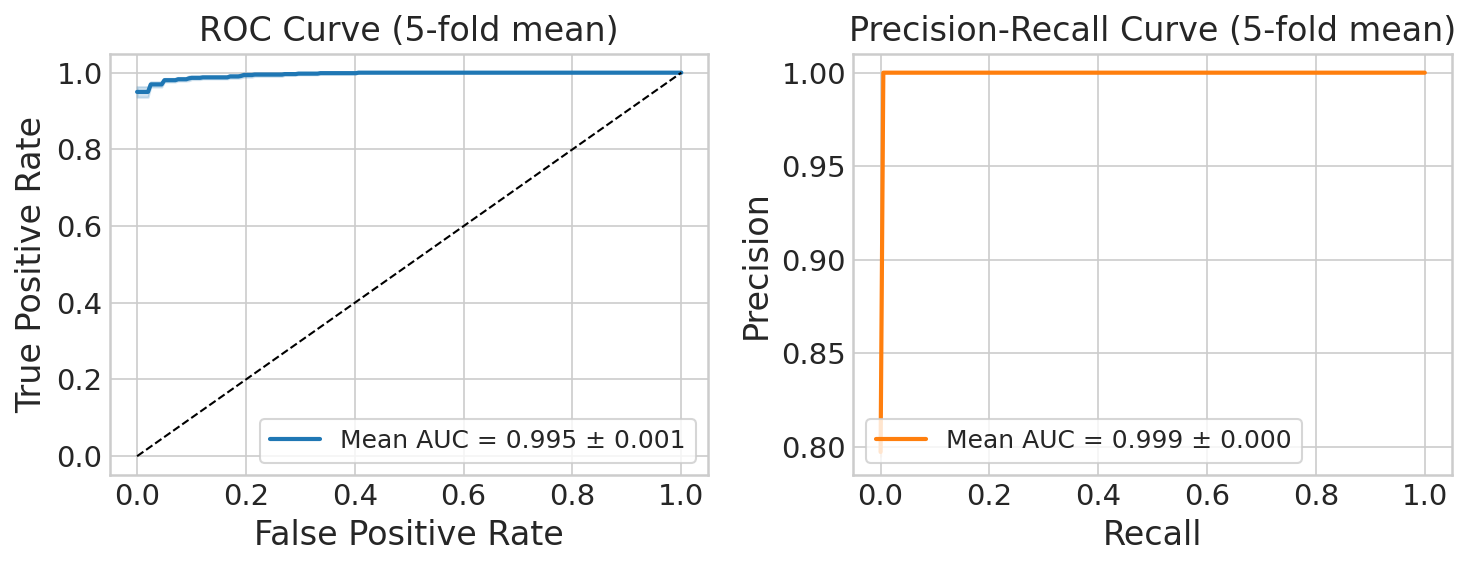}} 
    
    \subfloat[Pile CC]{\includegraphics[width=0.45\linewidth]{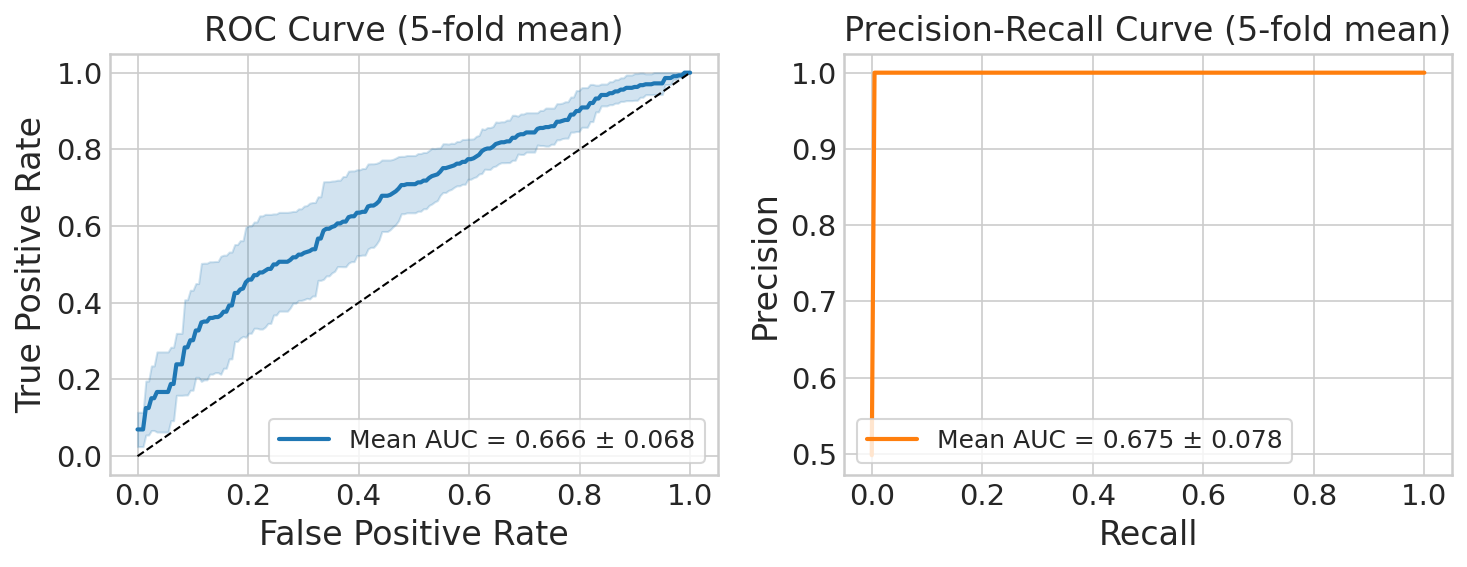}}
    \subfloat[PubMed Central]{\includegraphics[width=0.45\linewidth]{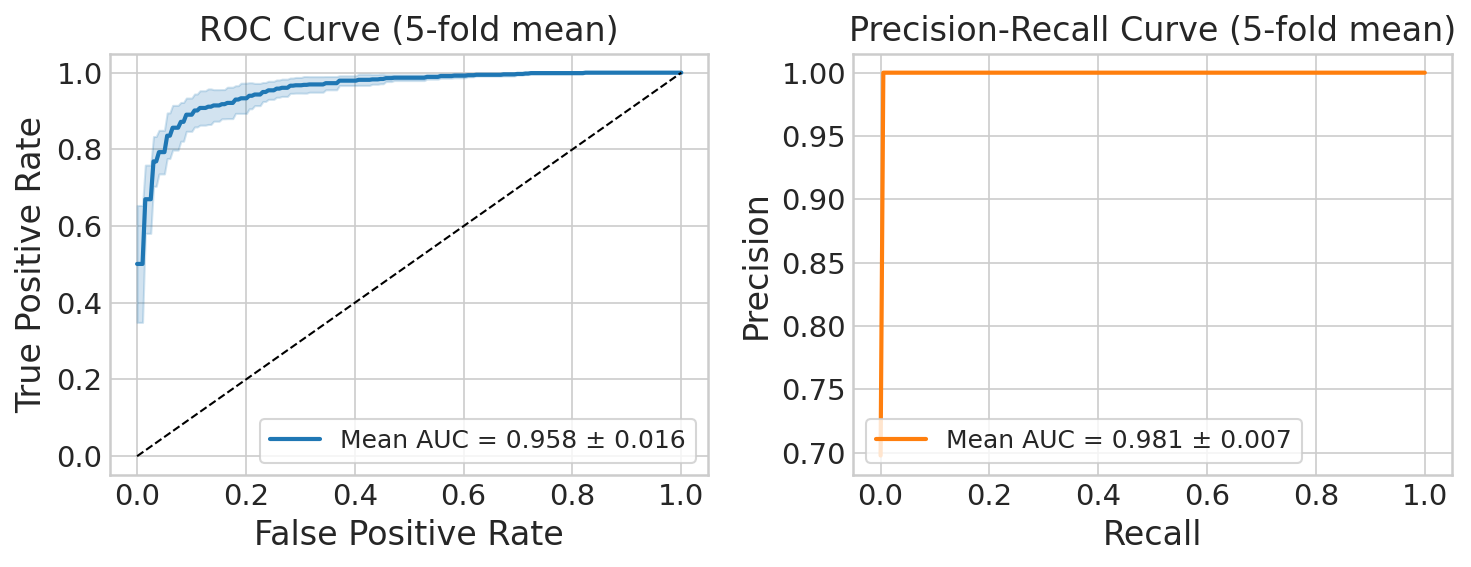}} \\ 

    \subfloat[Wikipedia]{\includegraphics[width=0.45\linewidth]{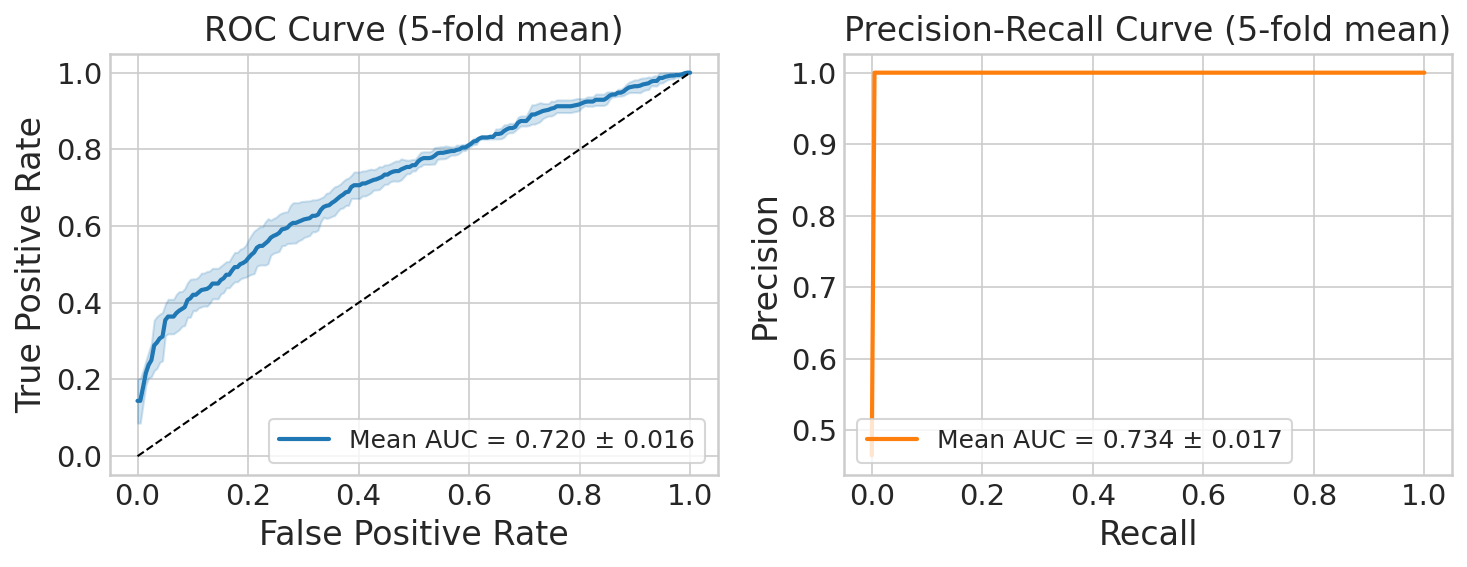}}
    \caption{ROC and Precision-Recall curves across 7 different MIMIR subsets and Pythia-1.4B model for AttenMIA. 
    Each subfigure shows the mean 5-fold ROC curve (left) and PR curve (right) for the corresponding subset. Shaded regions indicate standard deviation across folds.}
    \label{fig: ROC_AUC_MIMIR_Pythia-1.4B}
\end{figure*}

Figure~\ref{fig: ROC_AUC_MIMIR_Pythia-1.4B} presents the ROC and PR curves for the seven MIMIR subsets (\textit{arXiv}, \textit{DM Math}, \textit{GitHub}, \textit{HackerNews}, \textit{Pile CC}, \textit{PubMed Central}, and \textit{Wikipedia}) using the Pythia-1.4B model. Each subfigure shows the mean 5-fold ROC curve (left) and PR curve (right), with shaded regions indicating standard deviation across folds. 

The results confirm that \textit{AttenMIA} achieves strong discrimination between members and non-members across a wide range of operating points. In particular, subsets such as \textit{arXiv} and \textit{DM Math} exhibit near-perfect ROC and PR curves, reflecting high vulnerability to membership inference, whereas subsets like \textit{GitHub} and \textit{HackerNews} yield lower but still significant separability. Overall, these curves provide further evidence that attention-based features capture robust memorization signals across diverse domains.

